%% file: main.tex
\documentclass[10pt,twocolumn,letterpaper]{article}

\usepackage[pagenumbers]{cvpr}      

\input{preamble}

\definecolor{cvprblue}{rgb}{0.21,0.49,0.74}
\usepackage[pagebackref,breaklinks,colorlinks,citecolor=cvprblue]{hyperref}

\def\paperID{372} 
\def\confName{3DV\xspace}
\def\confYear{2025\xspace}

\title{MorphoSkel3D: Morphological Skeletonization of 3D Point Clouds for Informed Sampling in Object Classification and Retrieval}

\author{Pierre Onghena \qquad Santiago Velasco-Forero \qquad Beatriz Marcotegui\\
\\
Mines Paris, PSL University, Center for Mathematical Morphology (CMM), 77300 Fontainebleau, France\\
{\tt\small name.surname@minesparis.psl.eu}
}
\begin{document}
\maketitle
\input{sec/0_abstract}    
\input{sec/1_introduction}

\input{sec/2_related}
\input{sec/3_background}
\input{sec/4_method}
\input{sec/5_experiment}
\input{sec/6_conclusion}
\input{sec/7_acknowledgement}
{
    \small
    \bibliographystyle{ieeenat_fullname}
    \bibliography{main}
}
\input{sec/X_suppl}

\end{document}

%% file: preamble.tex
\usepackage[dvipsnames]{xcolor}

\newcommand{\realset}{\mathbb{R}}
\newcommand{\MedialAxis}{\mathcal{MA}}
\newcommand{\Occupancy}{\mathcal{O}}
\newcommand{\Mesh}{\mathcal{M}(T)}
\newcommand{\Closedsurface}{\mathcal{S}}
\newcommand{\Interiorvolume}{\mathcal{V}}
\newcommand{\Skeleton}{\mathcal{SK}}
\newcommand{\MorphoSkel}{\text{MS3D}}

\usepackage{float}
\usepackage{lipsum}
\usepackage{graphicx}
\usepackage{stfloats}
\usepackage{tabularx}
\usepackage{xstring}
\usepackage{multirow}
\usepackage{xspace}
\usepackage{pgf}
\usepackage{url}
\usepackage{subcaption}
\usepackage[hang,flushmargin]{footmisc}
\usepackage{makecell}
\usepackage{algorithm} 
\usepackage{algpseudocode}
\usepackage[super]{nth}
\usepackage{mathrsfs}
\usepackage{cancel}
\usepackage{times}
\usepackage{microtype}
\usepackage{epsfig}
\usepackage{caption}
\usepackage{float}
\usepackage{placeins}
\usepackage{color, colortbl}
\usepackage{stfloats}
\usepackage{enumitem}
\usepackage{comment}
\usepackage[inkscapelatex=false]{svg}

\DeclareMathOperator*{\argmin}{arg\,min}

%% file: sec/0_abstract.tex
\begin{abstract}
Point clouds are a set of data points in space to represent the 3D geometry of objects. A fundamental step in the processing is to identify a subset of points to represent the shape. While traditional sampling methods often ignore to incorporate geometrical information, recent developments in learning-based sampling models have achieved significant levels of performance. With the integration of geometrical priors, the ability to learn and preserve the underlying structure can be enhanced when sampling. To shed light into the shape, a qualitative skeleton serves as an effective descriptor to guide sampling for both local and global geometries. In this paper, we introduce MorphoSkel3D\footnote{Code: \url{https://github.com/Pierreoo/MorphoSkel3D}} as a new technique based on morphology to facilitate an efficient skeletonization of shapes. With its low computational cost, MorphoSkel3D is a unique, rule-based algorithm to benchmark its quality and performance on two large datasets, ModelNet and ShapeNet, under different sampling ratios. The results show that training with MorphoSkel3D leads to an informed and more accurate sampling in the practical application of object classification and point cloud retrieval.
\end{abstract}

%% file: sec/1_introduction.tex
\section{Introduction}
\label{sec:introduction}

A skeleton is a simplified representation of the surface of a 3D shape, which makes it essential in a prior analysis to understand how the surface points are connected at different levels, from coarse- to fine-grained parts. Other common shape descriptors, such as the centroid or volume, are limited to the formulation of global rather than local insights about the geometry. Extracting a skeleton, however, has proven to be a challenging task for 3D shapes due to its extensive complexity in geometric processing. More recently, deep learning-based, skeleton models have been introduced for point clouds in an unsupervised manner to estimate skeletal spheres. While the wide-spread use of deep learning has many benefits, it's also coupled with significant challenges. Whenever skeletons are derived from learnable models, it faces certain limitations despite its advanced capabilities. One drawback is the fixed generalization over shapes, as the network's performance is heavily dependent on the diversity of training data. This dependency can hinder skeleton estimation when encountering data that diverges from the training set. Usually, the model complexity also increases with the prediction of a large number of skeletal spheres such that the network struggles to include finer details. These requirements can be a barrier to widespread adoption and iterative experimentation.
\begin{figure}[t]
    \centering
    \includegraphics[width=0.5\textwidth]{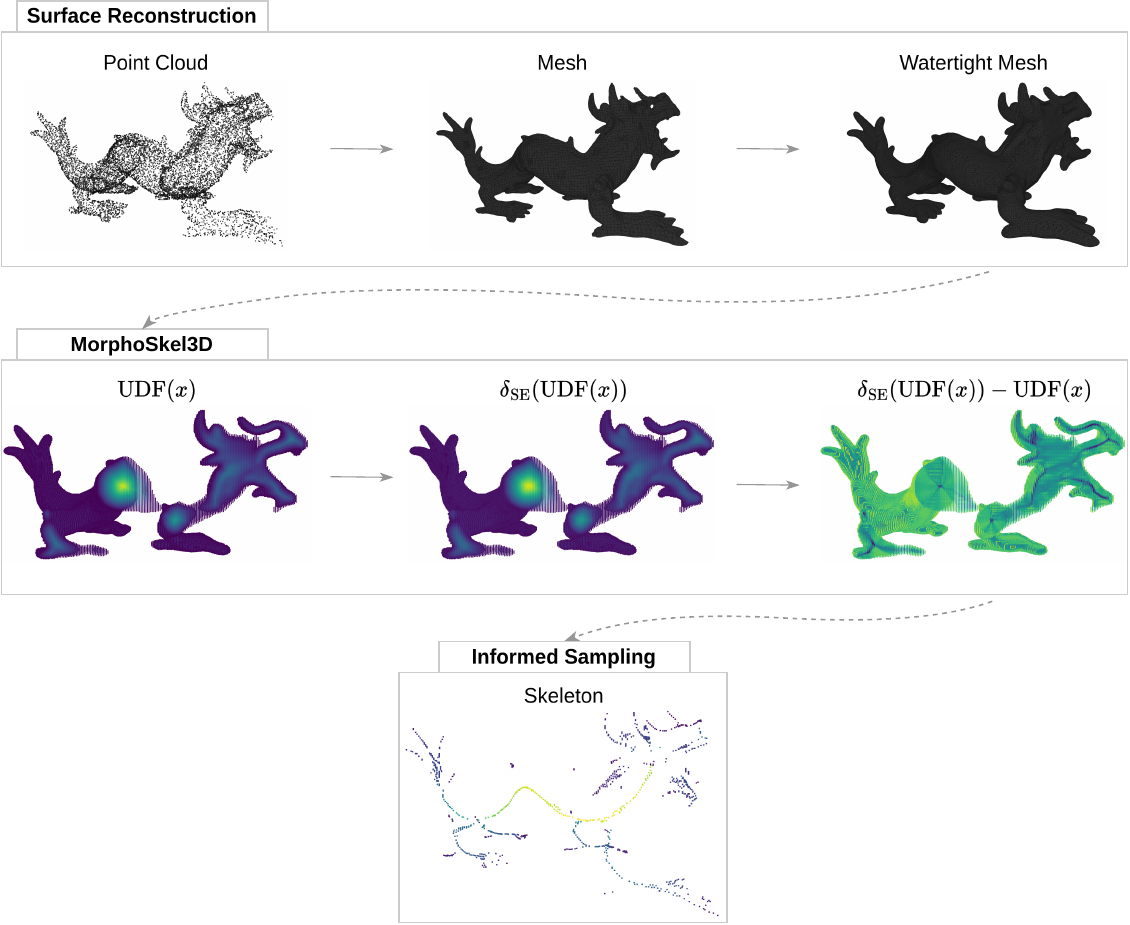}
    \caption{Overview of the proposed shape-agnostic skeletonization pipeline, illustrated with the Stanford dragon~\cite{10.1145/237170.237269}. In the MorphoSkel3D module, the local maxima on the unsigned distance function of inner points reveal the set of maximal balls. The local neighbourhood of points is encoded by a structuring element $\text{SE}$.}
    \label{fig:overview}
\end{figure}
 Moreover, it remains an open question how to formally define the set of skeletal spheres for 3D shapes. Replacing the skeletal network with the morphological operations depicted in the overview of \cref{fig:overview}, can alleviate these issues by providing a more consistent and efficient method. As morphology is not dependent on training data, it can accurately capture features with low computational demands. The contribution of this paper is threefold:

\begin{itemize}
    \item We propose a lightweight skeletonization method that is able to bring scalability to large-scale datasets.
    \item We integrate the skeleton as a prior to enable geometry-informed sampling of a representative subset.
    \item We extensively validate the skeletal quality of our method, demonstrating improved sampling performance in two tasks: object classification and point cloud retrieval.
\end{itemize}

%% file: sec/2_related.tex
\section{Related Work}
\label{sec:related}
Many research efforts in the field of computer vision have focused on the development of skeletonization techniques for 2D images. However, it has been more complex to extend a formal definition of skeletonization to the 3D space for point clouds to assist shape analysis and downstream tasks~\cite{tagliasacchi20163d}. This section discusses the concepts of skeletonization and morphology, to support its functionality for guided sampling.

\paragraph{3D shape skeletonization}
In recent years, several works have been proposed to tackle the sensitivity of the medial axis transform (MAT)~\cite{blum1967transformation} to irregular shape surfaces. For instance, Q-MAT~\cite{10.1145/2753755} / Q-MAT+~\cite{PAN201916} or Coverage Axis~\cite{dou2022coverage} / Coverage Axis++~\cite{wang2024coverage} developed an algorithm with hand-crafted local features to capture the geometric features that approximate a simplified medial axis transform. Whereas Q-MAT relies on watertight surfaces, Coverage Axis uses a winding numbers method~\cite{Barill:FW:2018} to select inner points for point cloud inputs. Learning-based methods learn a geometric transformation to mimic the properties of MAT, pioneered by Point2Skeleton~\cite{Lin_2021_CVPR}, the skeletal points are considered as a local center of surface points. Consequently, rather than directly predicting the skeletal points, a convex combination of input points and weights is learned through multiple loss functions to minimize the reconstruction error. Other methods implement a Laplacian-based contraction process~\cite{5521461, meyer2023cherrypicker}, or use the first Laplace-Beltrami eigenfunction to construct Reeb graphs for high level surface information, and the barycenters of these graphs are then connected to find the skeleton~\cite{Liang2012GeometricUO}. More recently, implicit neural representations have been introduced to provide continuous and detailed modeling of complex shapes~\cite{deluigi2023inr2vec}. For example, distance functions can be effectively fitted to obtain high resolution outputs of shapes~\cite{Park_2019_CVPR, chibane2020ndf}. Neural skeleton~\cite{CLEMOT2023368} leverages the implicit field function for a more accurate distance estimation near the medial axis to extract topological information. In order to provide a more formal definition, our method proposes to transfer the effectiveness of morphology for skeletonization from image analysis to the distance function of 3D point clouds.

\paragraph{3D morphology}
Research works in image processing have explored techniques that extend traditional 2D morphological operations, such as dilation and erosion, into the 3D image domain. These operations to form a skeleton have shown to be fundamental in preserving the topology of tabular structures in the 3D segmentation of medical images~\cite{cldice2021, Menten2023skeletonization}. To avoid the transformation of point clouds to a 2D or 3D image, morphological dilation and erosion operations have been proposed that focus on the addition and removal of points according to a structuring element~\cite{BALADO2020208}. Another framework, Point Morphology~\cite{CB:2014:PointMorphology}, defined a projection procedure of shifting points to find the operators that sample the medial axis in a meshless context. The extension of morphological operators to large point clouds has also been studied to better define boundaries for segmentation in urban scenes~\cite{10.1007/978-3-030-20867-7_29}. We suggest an alternative approach to implement morphological operators for efficient 3D point cloud skeletonization.

\paragraph{Point cloud sampling}
An obstacle in the processing of point clouds is the high density of data, which can hinder the effective use of deep learning models~\cite{landrieu:hal-01801186}. The aim of sampling is to extract a representative subset from the original point cloud to ensure that the resulting sampled point cloud is both representative of the original data and optimized for further processing and analysis. For instance, to maintain the classification and segmentation performances. This goal aligns with the method of Wen et al.~\cite{wen2023learnable} that explores the object skeleton to preserve geometrical information during sampling. In their work, the authors adopted a deep learning approach similar to Point2Skeleton~\cite{Lin_2021_CVPR} to estimate the skeleton. An ablation study between the learned skeleton and DPC~\cite{Dpoints15} skeleton demonstrated comparable results in their ability to improve the sampling of representative points. Therefore, as a metric to reflect skeletal quality, we incorporate our skeleton to evaluate its ability to inform sampling.


%% file: sec/3_background.tex
\section{Background}
\label{sec:background}

To sharpen the intuition before defining our method, this section reviews the background to afterwards introduce morphological skeletonization. The concept of a skeletal structure is first discussed as a geometric approach to introduce the notion of the skeleton of a set of inner points. The goal is to leverage the properties of a watertight mesh to accurately determine an occupancy and distance function of points lying within the object. Based on the distance of inner points to the surface, we propose a highly efficient process that is able to extract the skeletal points from a shape. The interest primarily lies within the inner points, as these serve as the candidate skeletal points. To achieve this, two sequential steps are employed. Firstly, an occupancy function is applied to determine the points within the mesh. Following this determination, an unsigned distance function is computed on these points to facilitate morphological operators to the extraction of the skeletal structure.

\subsection{Skeleton by maximal balls}
 A rich segment of mathematical literature is devoted to the study of the central part of a set in Euclidean space. While various definitions have been proposed, their central parts are often similar yet not always equivalent. In particular, given a closed surface $\Closedsurface$, the \emph{medial axis} $\MedialAxis(\Closedsurface)$ is identified by the set of centers of maximally inscribed spheres that are tangent to $\Closedsurface$ at two or more points. Usually referred to as the rowboat analogy, the medial axis has been introduced by Blum~\cite{blum1967transformation} in image analysis and by Beucher and Lantuejoul~\cite{BEUCHER1994127,lantuejoul1978squelettisation} in mathematical morphology. Other studies, including \cite{chaussard2011robust} and \cite{chazal2004stability}, study the medial axis from a different perspective. For any $x$ in volume $\Interiorvolume$ enclosed by $\Closedsurface$, let $D(x)$ denote the set of medial balls:
\begin{equation}
D(x)=\{y\in \Closedsurface :  d(x,y) = d(x,\Closedsurface)\},
\end{equation}
where $d(x,y)$ denotes the Euclidean distance between $x$ and $y$. The distance between the point and surface is defined as $d(x,\Closedsurface) = \inf_{y \in \Closedsurface} d(x,y)$, where $\inf$ denotes the infimum. Then, the medial axis of a closed surface $\Closedsurface$ is defined as:
\begin{equation}
\MedialAxis(\Closedsurface):=\{ x \in \Closedsurface : |D(x)|\geq 2 \},
\end{equation} 
where $|D(x)|$ denotes the cardinality of $D(x)$. The medial axis together with the distance from each point to its closest surface point defines the medial axis transform (MAT)~\cite{choi1997mathematical} as a powerful tool in computer vision and image processing. The related concept of a skeleton is often mistakenly conflated with the medial axis. The \emph{skeleton} of a closed surface $\Closedsurface$, denoted $\Skeleton(\Closedsurface)$, is namely defined as the set of centers of maximal balls inscribed in $\Closedsurface$. In specific, a ball $B(x, i)$ defined by its center at point $x$ and radius $i$ is considered maximal within $\Closedsurface$ if there exists no other ball $B(y, j)$ such that $B(x, i)$ is contained within $B(y, j)$ with a larger radius $j > i$. We emphasize that the set of centers of maximal inscribed disks is generally not connected in discrete distance maps~\cite{BEUCHER1994127,ge1996generation,couprie:hal-00622521}. Consequently, while this set is occasionally referred to as the skeleton, it is primarily useful for compression rather than for shape analysis tasks. A more conventional definition of a skeleton requires that it preserves the same simple connectivity of $\Closedsurface$, to make it homotopic to the surface. Typically, this involves a linkage of the centers  of maximal inscribed balls to produce a connected skeleton~\cite{ge1996generation,tagliasacchi20163d}. Although things can become more complicated in the 3D space, the proposed approach, that is lightweight and object-agnostic, is introduced in section \cref{sec:method}. The skeletal representation focuses on the maximal inscribed balls to provide a compact and meaningful abstraction of the original set.

\subsection{Occupancy Function}

\begin{figure}[ht!]
    \centering
    \begin{subfigure}{0.22\textwidth}
        \centering
        \includegraphics[width=\linewidth]{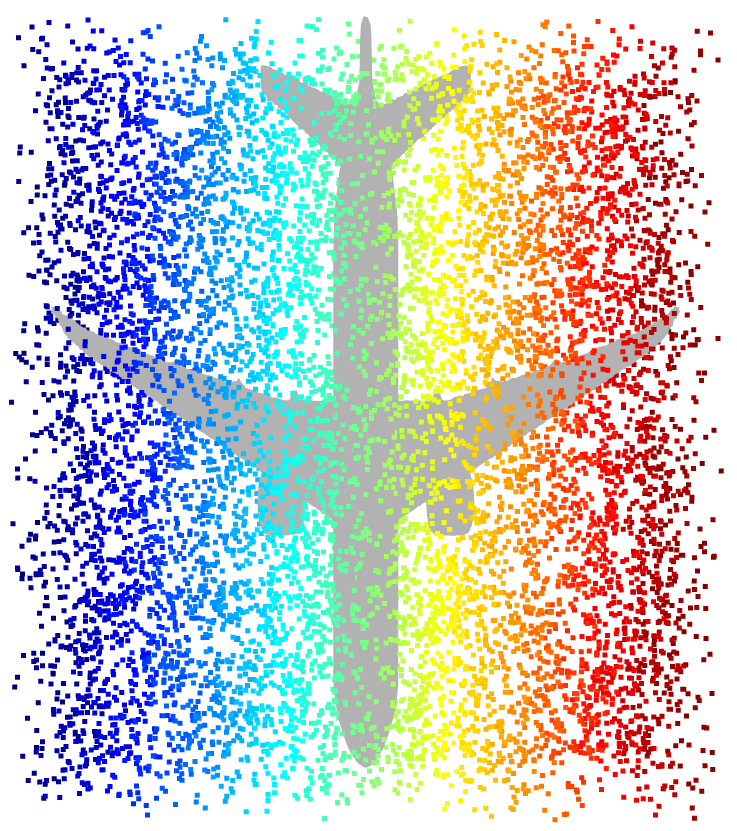}
        \label{fig:random}
    \end{subfigure}
    \hfill
    \begin{subfigure}{0.20\textwidth}
        \centering
        \includegraphics[width=\linewidth]{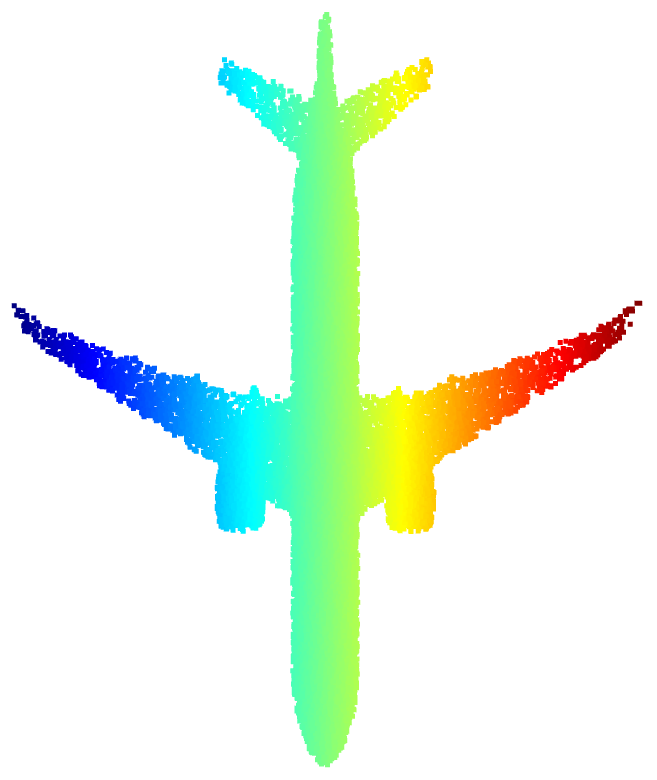}
        \\[0.6 cm]
        \label{fig:inner}
    \end{subfigure}
    \\[-0.4 cm]
    \caption{Occupancy function used to retain the inner points from the bounding box points. A case for 10000 points in the bounding box (left). Inner points extracted for the case with 1M points within the box volume (right).}
    \label{fig:occupancy}
\end{figure}

Based on a bounding box that encapsulates the object, random points are uniformly distributed in the volume to ensure the coverage of the entire shape. An example shape with bounding box points is illustrated on the left of \cref{fig:occupancy}. Following this, an occupancy function is applied to determine whether each bounding box point lies within the watertight surface mesh. It helps to identify the points inside the shape and excludes those outside its surface. The occupancy function that follows the ray casting algorithm~\cite{roth1982ray}, evaluates a given point by analyzing its intersections with the surface triangles. Let $p = (x, y, z) \in \realset^3$ be a bounding box point and $\Mesh$ the watertight mesh that consists of a set of triangles $T$. The number of intersections between the ray cast from $p$ and the triangles in $T$ is denoted by $N_{p}(T)$. This algorithm is also referred to as the crossing number algorithm or the even–odd rule algorithm~\cite{shimrat1962algorithm}. Then, the \emph{occupancy function} $\Occupancy: \realset^3 \mapsto \{0,1\}$ for the bounding box point $p$ with respect to the mesh $\Mesh$ can be computed by:
\begin{equation}
\begin{aligned}
\Occupancy(p) = 
\begin{cases} 
1, & \text{if } N_{p}(T) \text{ is odd}, \\
0, & \text{if } N_{p}(T) \text{ is even}.
\end{cases}
\end{aligned}
\label{eq:occupancy}
\end{equation}
If the number of intersections between a ray cast from $p$ and the triangles in $T$ is odd, $\Occupancy(p)$ equals to one, indicating that $p$ is inside the mesh. Conversely, if the number of intersections is even, $\Occupancy(p)$ equals to zero, indicating that $p$ is outside the mesh. An example illustrating the selection of inner points is provided on the right side of \cref{fig:occupancy}. By $\Occupancy$, we consider $\Mesh$ as a closed surface $\Closedsurface$ to ensure that the mesh forms a completely enclosed surface without gaps.

\subsection{Signed Distance Function}
Given a closed surface $\Closedsurface$ in metric space $\mathcal{C} \subseteq \realset^3$, a \emph{signed distance function} (SDF) is a continuous function that provides the closest distance from a given point $x \in \realset^3$ to the surface $\Closedsurface$. The sign indicates whether or not the point $x$ lies within the interior volume $\Interiorvolume$ enclosed by $\Closedsurface$. The signed distance function outputs a negative value for points inside the watertight mesh and increases as it approaches the surface or zero-level set. For points outside the mesh, it takes positive values that become larger as one moves further away from the surface:
\begin{equation}
\text{SDF}(x) = {
\begin{cases}
d(x,\Closedsurface) & {\text{if }} x \notin \Interiorvolume, \\
-d(x,\Closedsurface) & {\text{if }} x \in \Interiorvolume .
\end{cases}}
\label{eq:signed}
\end{equation}
The utilization of the occupancy function to discern points within a mesh offers an advantage to circumvent the necessity of performing the distance function for every point within the bounding box of the object. Having already identified the inner points, see \cref{fig:occupancy} right, the signed distance function could be simplified to the unsigned distance function. This approach eliminates the need to calculate distances for points outside the object's surface, thus optimizing computational efficiency. An example of the distance function is illustrated in \cref{fig:original}. Additionally, the \emph{unsigned distance function} (UDF) only provides positive distances such that further morphological operations can be simplified to:
\begin{equation}
\text{UDF}(x) = d(x,\Closedsurface) \quad {\text{for }} x \in \Interiorvolume.
\label{eq:unsigned}
\end{equation}

%% file: sec/4_method.tex
\section{Method} 
\label{sec:method}

The goal of Wen et al.~\cite{wen2023learnable} was to identify a subset of points that maintains classification performance using only a few selected points. This process involves the learning of a sampling strategy that is informed by prior knowledge of the skeleton. However, the skeleton estimation network is limited due to its inability to generalize to other shapes than the training set. In this work, we propose a novel skeletonization method, named MorphoSkel3D, that is object-agnostic through a series of morphological operators to give priority to the points belonging to the skeleton $\Skeleton(\Closedsurface)$.

\begin{figure*}[b]
    \centering
    \begin{subfigure}[b]{0.42\textwidth}
        \centering
        \includegraphics[width=\linewidth]{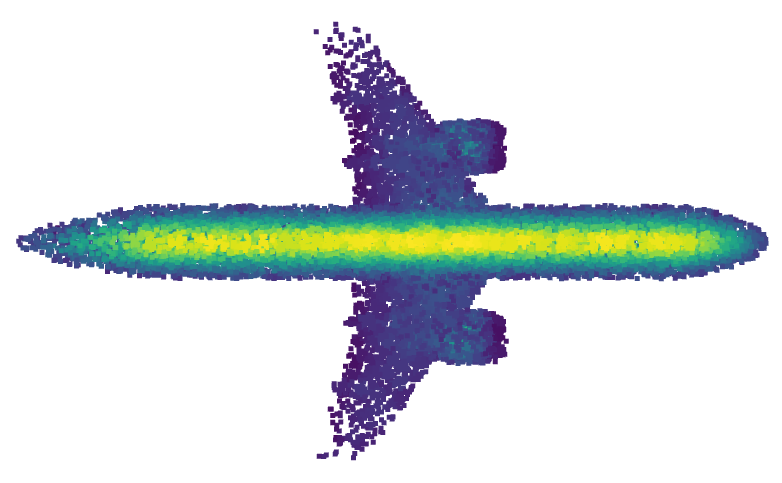}
        \caption{$\delta_{\text{SE}}(\text{UDF}(x))$}
        \label{fig:dilated}
    \end{subfigure}
    \hfill
    \begin{subfigure}[b]{0.42\textwidth}
        \centering
        \includegraphics[width=\linewidth]{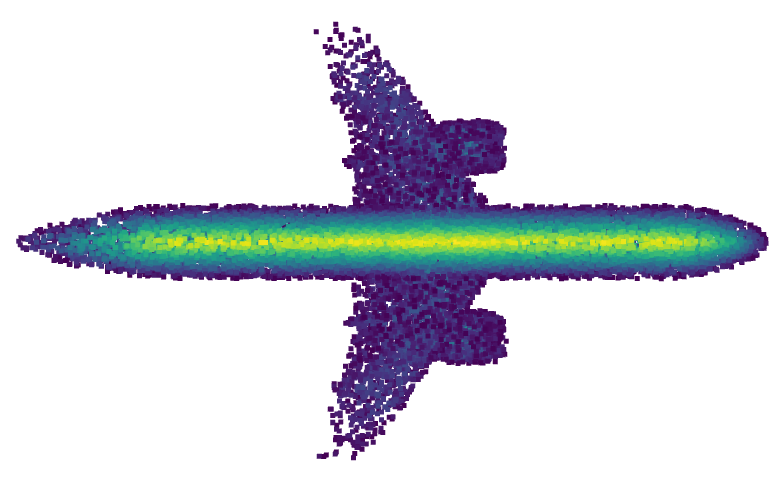}
        \caption{$\text{UDF}(x)$}
        \label{fig:original}
    \end{subfigure}
    \\[-1.0cm]
    \begin{subfigure}[b]{\textwidth}
        \centering
        \includegraphics[width=0.38\textwidth]{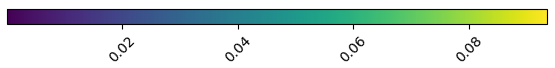}
    \end{subfigure}
\caption{Transformed distance function by dilation (left) and its initial unsigned distance function (right). Brighter points indicate greater distances, while darker areas correspond to points closer to the surface.}
\label{fig:morphological skeletonization}
\end{figure*}

\subsection{Morphological Skeletonization}
\begin{figure*}[t]
    \centering
    \begin{subfigure}[b]{0.42\textwidth}
        \centering
        \includegraphics[width=0.97\textwidth]{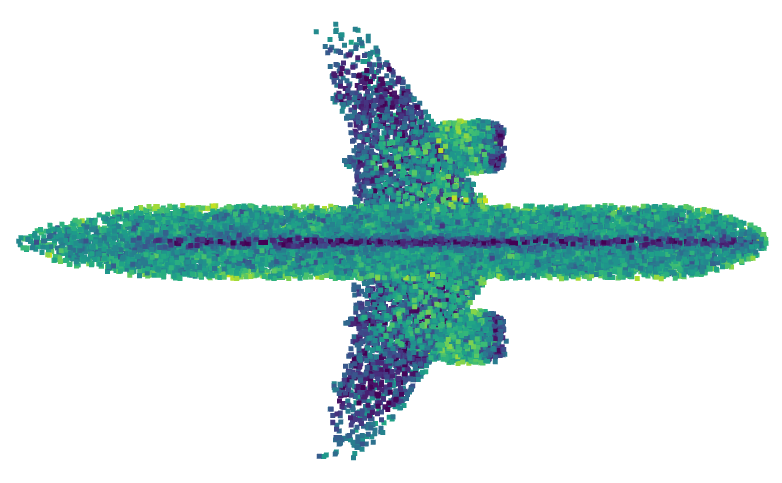}
        \includegraphics[width=0.90\textwidth]{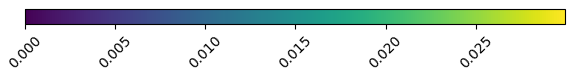}
        \caption{$\MorphoSkel(x)$}
        \label{fig:difference}
    \end{subfigure}
    \hfill
    \begin{subfigure}[b]{0.42\textwidth}
        \centering
        \includegraphics[width=0.9\textwidth]{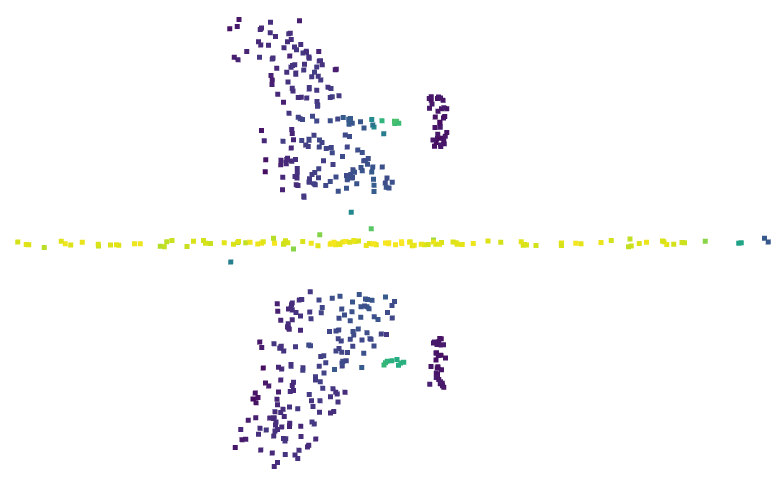}
        \\[0.25cm]
        \includegraphics[width=0.90\textwidth]{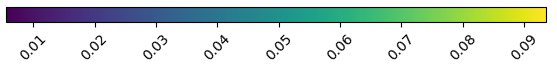}
        \caption{1024 skeletal points with lowest $\MorphoSkel(x)$}
        \label{fig:retained skeleton}
    \end{subfigure}
\caption{Based on 1M bounding box points, $\MorphoSkel(x)$ as the difference between the dilated and original UDF (left). The 1024 inner points with lowest $\MorphoSkel(x)$ values are retained to form the skeleton with point colors based on their original UDF to the surface (right).}
\label{fig:morphological skeletonization2}
\end{figure*}

In this method, a sequence of concepts and operators is presented based on the notion of distance. The objective is to outline a technique that characterizes a set of points based on the considerations of metric and geometry. An important concept includes that of a skeleton, which is of fundamental importance in computer vision, and has many practical applications for point clouds. In the analysis of the unsigned distance function, a property becomes transparent that local maxima correspond to maximal balls within the given space. These local maxima represent regions where the distance function reaches its greatest value within a neighborhood. The identification of these maxima becomes essential for understanding the structure of space. An approach to tackle this involves the study of the difference between the original distance and its \emph{dilation}. This analysis serves as a means to identify local maxima and mark the arrangement of a skeleton. Hence, the difference in the distance function among the inner points of $\mathcal{S}$ is described as:
\begin{equation}
\MorphoSkel(x) = \delta_{\text{SE}}(\text{UDF}(x)) - \text{UDF}(x) \quad {\text{for }} x \in \mathcal{V},
\label{eq:skeleton}
\end{equation}
where $\delta_{\text{SE}}$ denotes a dilation equipped by the structuring element $\text{SE}$. A dilation operator takes the maximum value of $\text{UDF}$ within the neighborhood defined by $\text{SE}$. Note that Rosenfeld and Pfaltz~\cite{rosenfeld1966sequential} have shown for 2D images that, to find the set of maximal balls, it is sufficient to detect the local maxima on the distance function. With a graph to construct local connections, we use the same idea for 3D shapes to highlight the inner points that are candidates to be maximal ball centers. An option to define the structuring element is by using a $k$-nearest neighbor ($k$-NN) graph. For a point $x \in \mathcal{V}$, a directed graph $G = (V, E)$ is constructed, representing the local structure of the volume point cloud, where $V$ and $E$ denote the vertices and edges, respectively. The structuring element $\text{SE}$ is then defined as:
\begin{equation}
\text{SE}(x) = k\text{-NN}(x) \quad {\text{for }} x \in \mathcal{V}.
\label{eq:knn}
\end{equation}
The local information encapsulated in this graph is leveraged to take the maximum value of the distance function in the neighborhood. We denote this operation of transforming the distance function by dilation as $\delta_{\text{SE}}$. An illustration of a dilated distance function is presented in \cref{fig:dilated}, next to the initial unsigned distance function on the right in \cref{fig:original}. The resulting $\MorphoSkel(x)$ is illustrated in \cref{fig:difference} to provide a visual understanding of the transformation between the dilation operation and the original distance function.


\begin{figure}[ht!]
  \centering
  \begin{minipage}[b]{0.495\linewidth}
    \centering
    \includegraphics[width=\linewidth]{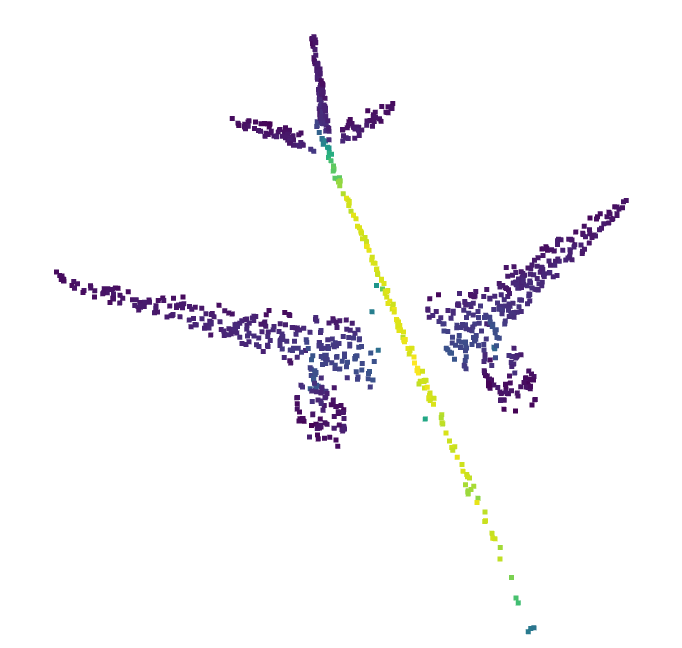}
    \includegraphics[width=0.85\linewidth]{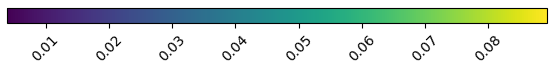}
  \end{minipage}
  \begin{minipage}[b]{0.495\linewidth}
    \centering
    \includegraphics[width=\linewidth]{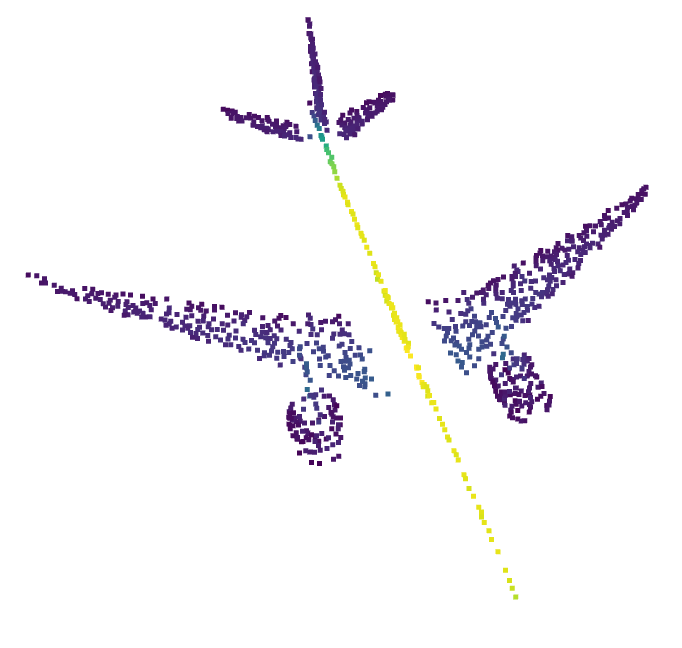}
    \includegraphics[width=0.85\linewidth]{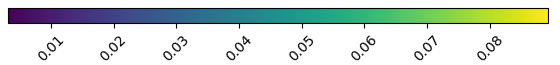}
  \end{minipage}
  \\[0.3cm]
  \begin{minipage}[b]{0.495\linewidth}
    \includegraphics[width=\linewidth]{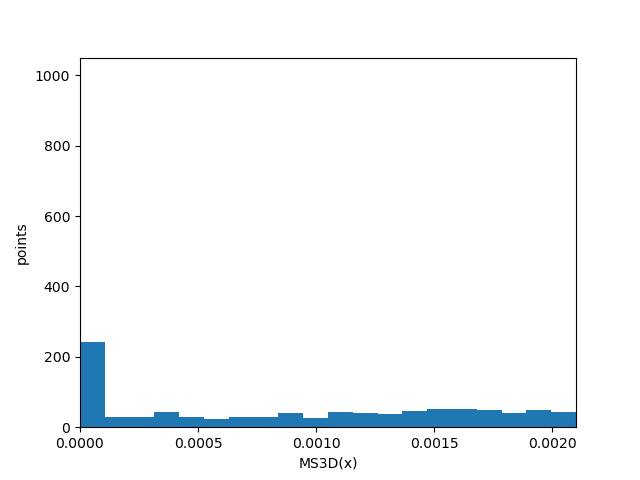}
  \end{minipage}
  \begin{minipage}[b]{0.495\linewidth}
    \includegraphics[width=\linewidth]{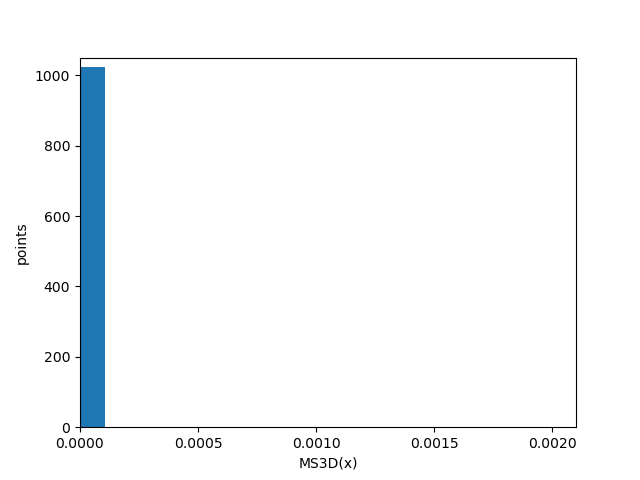}
  \end{minipage}
  \caption{The 1024 inner points selected by $\MorphoSkel(x)$, color-coded by their original UDF to the surface (up). Distribution of $\MorphoSkel(x)$ for 1024 skeletal points (down), contingent to the density of 1M (left) to 10M (right) bounding box points.}
  \label{fig:four_plots}
\end{figure}

To make sure that every skeleton used for training has the same number of points, a subset $\mathcal{K}$ from the set of inner points $\mathcal{V}$ is selected with the lowest $\MorphoSkel(x)$ values, where $\MorphoSkel(x)$ measures how close a point is to being a true skeletal point. We consider that the skeletal points, by definition of maximal balls, have a $\MorphoSkel(x)$ value equal to zero. It involves identifying a subset $\mathcal{K}$ of size $n$ that minimizes the sum of $\MorphoSkel(k)$ for all elements $k \in \mathcal{K}$:
\begin{equation}
\argmin_{\mathcal{K} \subset \{1,2,\ldots, S\}: |\mathcal{K}| = n} \sum_{k \in \mathcal{K}} \MorphoSkel(k).
\end{equation}
As a result, $\mathcal{K}$ consists of points closest to the center of a maximal ball and defines the skeletal points by minimizing $\MorphoSkel(x)$ values to provide a consistent subset for network training. In \cref{fig:four_plots}, the distribution of $\MorphoSkel(x)$ values reveals its dependency on the number of bounding box points. As the number of points in the bounding box increases, more accurate skeletal points appear. Although the skeleton becomes better defined, the increase in bounding box lightly prolongs the computational time required for its creation. The \cref{fig:retained skeleton} illustrates a subset of skeletal points retained by our proposed method. It shows that the inner points with lowest $\MorphoSkel(x)$ values are used to find the maximal balls that form the skeleton. The skeletal points in this representation are color-coded based on their initial UDF values to the surface to represent skeletal spheres.

\subsection{Prior Sampling Probability}

\begin{figure}[ht!]
    \centering
    \includegraphics[width=0.47\textwidth]{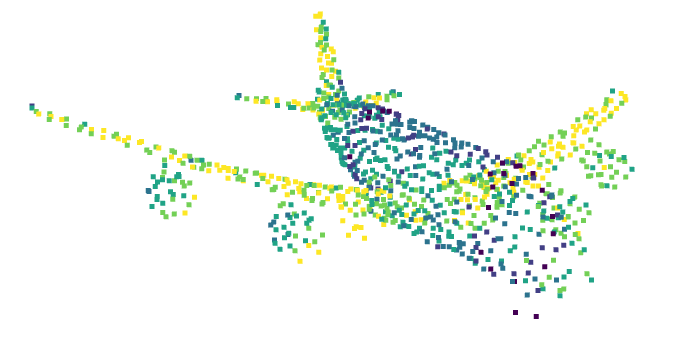}
    \includegraphics[width=0.33\textwidth]{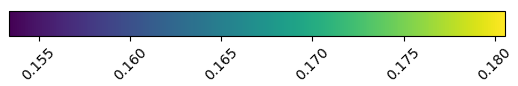}
    \caption{Heatmap of the prior per-point sampling weight, derived from the proposed MS3D skeletal structure.}
    \label{fig:heatmap}
\end{figure}

 The sampling process can be informed by prior knowledge that sits contained in the object's skeleton. By utilizing the concept of \emph{local feature size} (LFS), which measures the Euclidean distance from a surface point $p \in \mathcal{M}$ to the medial axis or skeleton, it becomes apparent that smaller LFS values denote intricate regions in the object. Hence, the strategy involves an augmentation of sampling probability in small LFS areas to enhance the overall representation. A heatmap where the prior sampling probability is visualized for each surface point is shown in \cref{fig:heatmap}. In this heatmap, areas with higher weights correspond more to fine-grained regions. In this way, the overall shape will be represented in the subset to improve the task network in its practical application.

%% file: sec/5_experiment.tex
\section{Experiments} 
\label{sec:experiments}

To evaluate the quality of the skeleton through morphology, the experimental setup is firstly discussed to assess the skeleton extraction in terms of reconstruction error and processing time. Once these metrics are compared, we apply our skeleton on two downstream tasks: object classification and point cloud retrieval.

\subsection{Experimental Setup}

\paragraph{Datasets}
The ShapeNet~\cite{shapenet2015} and ModelNet~\cite{DBLP:journals/corr/WuSKTX14} dataset are employed to establish an evaluation about the skeleton and its functionality as a prior. To compare our morphological skeleton, we use the same subset of the ShapeNet dataset as the Point2Skeleton~\cite{Lin_2021_CVPR} benchmark that includes 7088 shapes from eight categories. Each object has 2000 points for surface reconstruction to allow the extraction of its mesh and the selection of inner points. For the downstream tasks enhanced by a skeletal prior, the normalized and resampled ModelNet40 dataset of 12311 point clouds spread over 40 categories is used to align the benchmark with existing methods. In this dataset, each shape has 10000 points to reconstruct its surface that is made watertight to determine the inner bounding box points. The skeletons for both datasets are in turn extracted from the inner points by our MorphoSkel3D method. Based on the analysis in \cref{fig:four_plots}, the number of bounding box points is set to 10M to optimize the quality of the skeleton while keeping a low computation time. The resulting dataset consists of 1024 skeletal spheres that are ordered in ascending $\MorphoSkel(x)$ value. We notice that this setting includes accurate skeletal points with $\MorphoSkel(x)$ values of zero or close to.

\paragraph{Architectures}
For the first downstream task of object classification, Wen et al.~\cite{wen2023learnable} propose to learn a representative subset of the original point cloud with a selection method that is trained with the Gumbel-softmax trick to enable discrete yet differentiable sampling:
\begin{equation}
\mathcal{P}_{\text{sub}} = [Gs_{\tau} (Lp(\mathcal{W}_{\text{LFS}} \odot \mathcal{W}_{\theta})]^\top \mathcal{P},
\end{equation}
where $\mathcal{P}_{\text{sub}} \in \mathbb{R}^{M \times 3}$ denotes the extracted subset derived from the input point cloud $\mathcal{P} \in \mathbb{R}^{N \times 3}$. The weight $\mathcal{W}_{\text{LFS}} \in \mathbb{R}^{N \times 1}$ is computed in the prior phase from the skeleton, while $\mathcal{W}_{\theta} \in \mathbb{R}^{N \times D_{\text{sp}}}$ represents the feature matrix obtained from the sampling network, with $D_{\text{sp}}$ denoting the feature dimension. The linear mapping layer $Lp(\cdot) : \mathbb{R}^{N \times D_{\text{sp}}} \to \mathbb{R}^{N \times M}$ is applied, and the element-wise Hadamard product operation is denoted as $\odot$. The $Gs_{\tau} (\cdot)$ operation refers to the Gumbel-softmax where the confidence of the softmax is controlled by an annealed temperature $\tau$. For the sampling network to obtain a feature matrix $\mathcal{W}_{\theta}$, the DGCNN~\cite{dgcnn} architecture is applied with four edge convolution layers to capture local geometric features between points in the dynamic feature graph. The learned embedding $\mathcal{W}_{\theta}$ is merged with the skeletal information $\mathcal{W}_{\text{LFS}}$ by Hamard product to facilitate the Gumbel-softmax to learn the optimal subset of points for the task. During the sampling process, a pre-trained PointNet~\cite{qi2016pointnet} classification model operates in evaluation mode with its loss to guide the sampling network in classifying objects on fewer points. For the second task of point cloud retrieval, the trained sampling model of the first task is in evaluation to sample points for the classification model to form a global descriptor and compare representations for retrieval.

\paragraph{Metrics}
We follow other skeletonization techniques in the evaluation to ensure a fair comparison. Namely, the reconstruction error is evaluated to quantify how accurately the generated skeletal spheres reconstruct the original point cloud. Another way to evaluate the skeleton is by comparing it to the manually simplified MAT, which serves as the ground truth skeleton in the Point2Skeleton~\cite{Lin_2021_CVPR} benchmark. This dual evaluation approach uses the Chamfer and Hausdorff distance to provide a reflection of discrepancy between the two sets. However, relying solely on these metrics does not fully capture the method's skeletonization capability, especially when  a small reconstruction error is returned for a generated skeleton that closely matches the surface point cloud. Therefore, a benchmark on two downstream applications is also conducted to measure the skeleton's ability to improve sampling. For the classification task, the evaluation metric is based on the overall accuracy (OA) across all categories. The evaluation scheme for point cloud retrieval aligns with the mean Average Precision (mAP) that is derived from averaging the precision scores across point clouds based on $L_2$ distances between shape descriptors.

\subsection{Skeleton Extraction}

\begin{table*}[t]
\centering
\resizebox{\textwidth}{!}{%
\begin{tabular}{c|cccc|cccc|cccc|cccc}
\Xhline{1.5pt}
& \multicolumn{4}{c|}{CD-Recon} & \multicolumn{4}{c|}{HD-Recon} & \multicolumn{4}{c|}{CD-MAT} & \multicolumn{4}{c}{HD-MAT} \\
\Xhline{1.5pt}
& L1 & DPC & P2S & \cellcolor{gray!20}MS3D & L1 & DPC & P2S & \cellcolor{gray!20}\MorphoSkel & L1 & DPC & P2S & \cellcolor{gray!20}\MorphoSkel & L1 & DPC & P2S & \cellcolor{gray!20}\MorphoSkel\\
\Xhline{1.5pt}
Airplane & 0.0378 & 0.0348 & 0.0363 & \cellcolor{gray!20}\textbf{0.0152} & 0.2216 & 0.1436 & 0.1266 & \cellcolor{gray!20}\textbf{0.0933} & 0.0793 & 0.1307 & 0.0611 & \cellcolor{gray!20}\textbf{0.0368} & 0.2384 & 0.2580 & 0.1721 & \cellcolor{gray!20}\textbf{0.1598} \\
Chair & 0.1126 & 0.0769 & 0.0441 & \cellcolor{gray!20}\textbf{0.0296} & 0.4810 & 0.2478 & \textbf{0.1618} & \cellcolor{gray!20}0.2347 & 0.1885 & 0.2286 & 0.0974 & \cellcolor{gray!20}\textbf{0.0659} & 0.4991 & 0.3707 & \textbf{0.2151} & \cellcolor{gray!20}0.3123 \\
Table & 0.1041 & 0.0853 & 0.0424 & \cellcolor{gray!20}\textbf{0.0366} & 0.3453 & 0.2584 & \textbf{0.1745} & \cellcolor{gray!20}0.2205 & 0.1541 & 0.2683 & 0.0876 & \cellcolor{gray!20}\textbf{0.0823} & 0.3583 & 0.3690 & \textbf{0.2085} & \cellcolor{gray!20}0.3107 \\
Lamp & 0.1542 & 0.0712 & 0.0335 & \cellcolor{gray!20}\textbf{0.0265} & 0.3956 & 0.1850 & \textbf{0.1382} & \cellcolor{gray!20}0.1896 & 0.1870 & 0.1751 & 0.0884 & \cellcolor{gray!20}\textbf{0.0606} & 0.4089 & 0.2627 & \textbf{0.2003} & \cellcolor{gray!20}0.2595 \\
Guitar & 0.0655 & 0.0212 & 0.0179 & \cellcolor{gray!20}\textbf{0.0085} & 0.2180 & 0.0589 & 0.0625 & \cellcolor{gray!20}\textbf{0.0553} & 0.0817 & 0.0672 & 0.0536 & \cellcolor{gray!20}\textbf{0.0259} & 0.2262 & 0.1226 & 0.1216 & \cellcolor{gray!20}\textbf{0.1046} \\
Earphone & 0.0437 & 0.0573 & 0.0399 & \cellcolor{gray!20}\textbf{0.0245} & 0.1908 & 0.2059 & \textbf{0.1125} & \cellcolor{gray!20}0.1686 & 0.0607 & 0.2216 & 0.1638 & \cellcolor{gray!20}\textbf{0.0934} & \textbf{0.1732} & 0.3403 & 0.2130 & \cellcolor{gray!20}0.2815 \\
Mug & 0.2864 & 0.1280 & \textbf{0.0417} & \cellcolor{gray!20}0.0580 & 0.9142 & 0.3510 & \textbf{0.1419} & \cellcolor{gray!20}0.3201 & 0.5316 & 0.4600 & 0.1179 & \cellcolor{gray!20}\textbf{0.1126} & 0.9057 & 0.4308 & \textbf{0.2158} & \cellcolor{gray!20}0.4121 \\
Rifle & 0.0260 & 0.0215 & 0.0213 & \cellcolor{gray!20}\textbf{0.0097} & 0.1078 & 0.0702 & 0.0767 & \cellcolor{gray!20}\textbf{0.0584} & 0.0494 & 0.0427 & 0.0356 & \cellcolor{gray!20}\textbf{0.0250} & 0.1234 & 0.1050 & 0.0957 & \cellcolor{gray!20}\textbf{0.0873} \\
\Xhline{1.5pt}
Average & 0.1038 & 0.0668 & 0.0372 & \cellcolor{gray!20}\textbf{0.0274} & 0.3593 & 0.2049 & \textbf{0.1424} & \cellcolor{gray!20}0.1857 & 0.1665 & 0.2026 & 0.0828 & \cellcolor{gray!20}\textbf{0.0629} & 0.3667 & 0.3047 & \textbf{0.1898} & \cellcolor{gray!20}0.2601 \\
\Xhline{1.5pt}
\end{tabular}
}
\caption{Comparison of reconstruction error between the skeletal spheres of skeletonization methods to the surface point cloud (Recon) and the ground truth skeleton (MAT), Chamfer (CD) and Hausdorff (HD) distances.}
\label{tab:reconstructionresults}
\end{table*}

\begin{figure}[ht!]
    \centering
    \begin{subfigure}[b]{0.235\textwidth}
        \centering
        \includegraphics[width=0.99\linewidth]{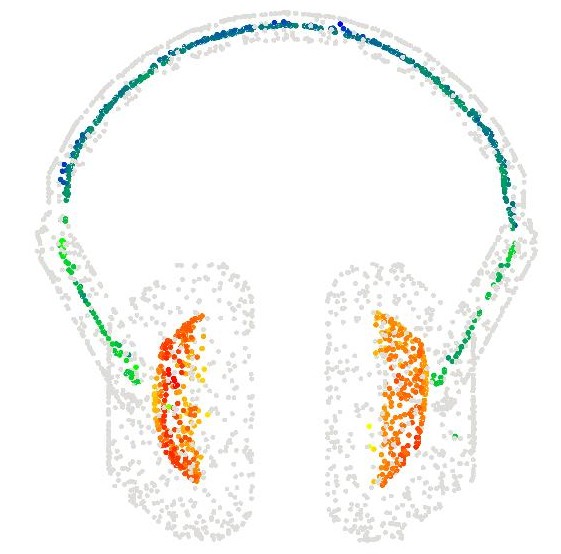}
    \end{subfigure}
    \hfill
    \begin{subfigure}[b]{0.235\textwidth}
        \centering
        \includegraphics[width=0.99\linewidth]{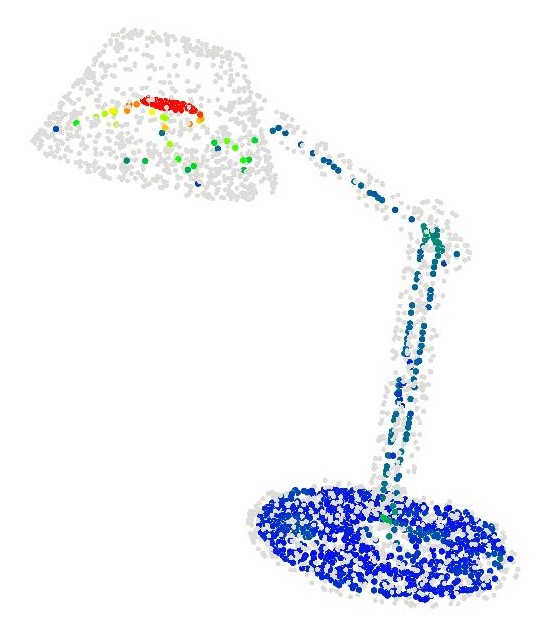}
    \end{subfigure}
\caption{The skeletal spheres of MS3D together with its surface points for an earphone and lamp example of the ShapeNet subset.}
\label{fig:shapenetvisualization}
\end{figure}

\begin{figure}[ht!]
    \centering
    \begin{subfigure}[b]{0.25\textwidth}
        \centering
        \includegraphics[width=0.99\linewidth]{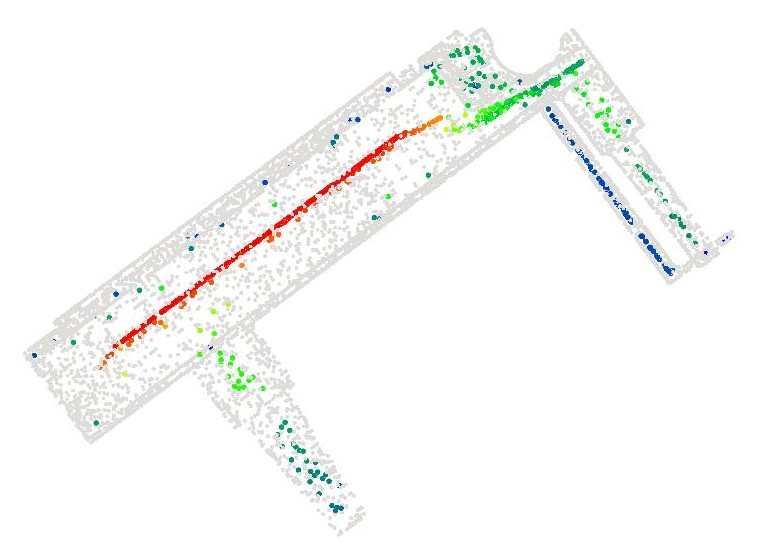}
    \end{subfigure}
    \hfill
    \begin{subfigure}[b]{0.22\textwidth}
        \centering
        \includegraphics[width=0.99\linewidth]{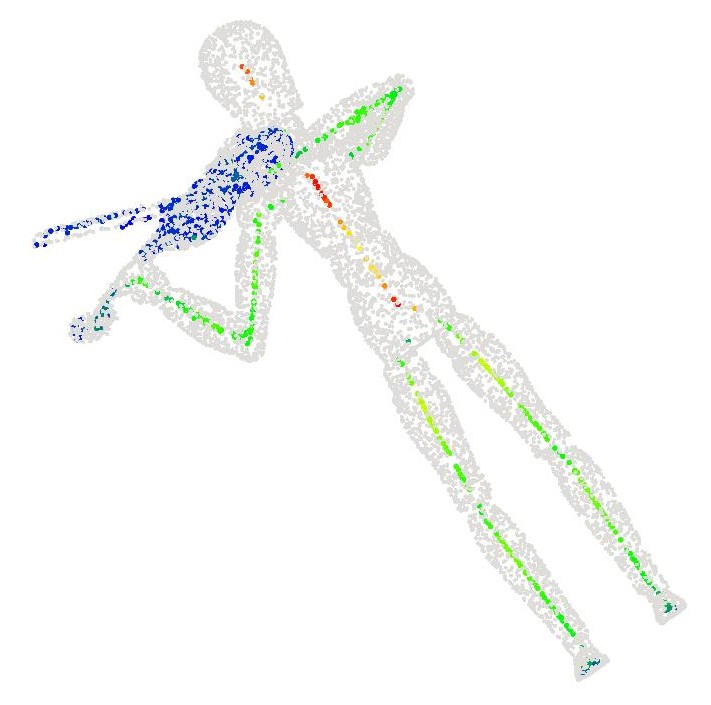}
    \end{subfigure}
\caption{The skeletal spheres of MS3D together with its surface points for a piano and person example of the ModelNet dataset.}
\label{fig:modelnetvisualization}
\end{figure}

The initial step before skeletonization is to convert a surface point cloud into its mesh representation with a shape reconstruction method. Among traditional methods, we opt for the Poisson surface reconstruction~\cite{Kazhdan:2006:PSR:1281957.1281965} algorithm to transform the test objects of the ShapeNet subset into meshes. Consequently, we generate watertight meshes~\cite{huang2018robust} in the surface reconstruction module to compute a distance function for the inner points. The $k$ is set to 20 for the $k$-NN graph to perform a dilation over the inner points in \cref{eq:skeleton}. In the skeletal analysis, we compare against other methods as $L_{1}$-medial skeleton~\cite{10.1145/2461912.2461913}, deep point consolidation (DPC)~\cite{Dpoints15}, and the learning-based Point2Skeleton (P2S)~\cite{Lin_2021_CVPR}. Whereas the number of estimated skeletal spheres is 100 for P2S, we select 1024 points to form the set of maximal balls for MS3D. For two examples of ShapeNet, a qualitative visualization of our method is given in \cref{fig:shapenetvisualization} to illustrate the skeletal spheres in relation to the original point cloud. The reconstruction results are reported in \cref{tab:reconstructionresults}. We observe that MS3D achieves the lowest Chamfer distance to prove its ability to define maximal balls and effectively extract skeletal spheres. A lower Hausdorff distance of MS3D is observed in three out of eight categories, but the other categories reveal its dependence on a surface reconstruction method to provide reliable meshes. Namely, as reconstruction is not evident with 2000 surface points, certain areas could have incorrect reconstruction to include false inner points for MS3D to skeletonize. Therefore, we implement PoNQ~\cite{maruani2024ponqneuralqembasedmesh} as a state-of-the-art optimization-based reconstruction method to provide meshes for the downstream tasks with the ModelNet dataset. Two examples of ModelNet are presented in \cref{fig:modelnetvisualization} to display the robust, object-agnostic skeletonization without the need to learn a geometric transformation. The corresponding processing times from surface reconstruction to the MS3D module are reported in \cref{tab:processing} for each category to highlight it's efficiency and steadiness on an Intel(R) Core(TM) i7-12800H CPU with 32 GB memory. For reference, the processing time to skeletonize with $L_1$ or DPC on a 100K point cloud is unsteady but takes approximately 60 seconds next to the high Chamfer and Hausdorff distances.

\begin{table}[ht!]
\centering
\resizebox{0.48\textwidth}{!}{%
\begin{tabular}{c|c|cc|cc|cc}
\Xhline{1.5pt}
& & \multicolumn{2}{c|}{Poisson~\cite{Kazhdan:2006:PSR:1281957.1281965}} & \multicolumn{2}{c|}{Watertight~\cite{huang2018robust}} & \multicolumn{2}{c}{\cellcolor{gray!20}\MorphoSkel} \\
\Xhline{1.5pt}
& Num & $T_{Total}$ & $T_{Average}$ & $T_{Total}$ & $T_{Average}$ & \cellcolor{gray!20}$T_{Total}$ & \cellcolor{gray!20}$T_{Average}$ \\
\Xhline{1.5pt}
Airplane & 100 & 34.83 & 0.35 & 119.79 & 1.20 & \cellcolor{gray!20}73.99 & \cellcolor{gray!20}0.74 \\
Chair & 334 & 98.57 & 0.30 & 590.40 & 1.70 & \cellcolor{gray!20}538.77 & \cellcolor{gray!20}1.61 \\
Table & 334 & 103.68 & 0.31 & 559.71 & 1.68 & \cellcolor{gray!20}661.61 & \cellcolor{gray!20}1.98 \\
Lamp & 167 & 53.04 & 0.32 & 282.66 & 1.69 & \cellcolor{gray!20}273.76 & \cellcolor{gray!20}1.64 \\
Guitar & 100 & 28.66 & 0.29 & 190.81 & 1.91 & \cellcolor{gray!20}144.97 & \cellcolor{gray!20}1.45 \\
Earphone & 13 & 3.92 & 0.30 & 21.47 & 1.65 & \cellcolor{gray!20}17.55 & \cellcolor{gray!20}1.35 \\
Mug & 36 & 10.20 & 0.28 & 39.01 & 1.08 & \cellcolor{gray!20}105.14 & \cellcolor{gray!20}2.92 \\
Rifle & 100 & 33.57 & 0.34 & 147.12 & 1.47 & \cellcolor{gray!20}160.25 & \cellcolor{gray!20}1.60 \\
\Xhline{1.5pt}
\end{tabular}
}
\caption{Comparison of total and average processing times across categories from surface reconstruction to MorphoSkel3D, time (s).}
\label{tab:processing}
\end{table}

\subsection{Object Classification}
\begin{figure*}[t]
    \centering
    \begin{subfigure}[b]{0.245\textwidth}
        \centering
        \includegraphics[width=\linewidth]{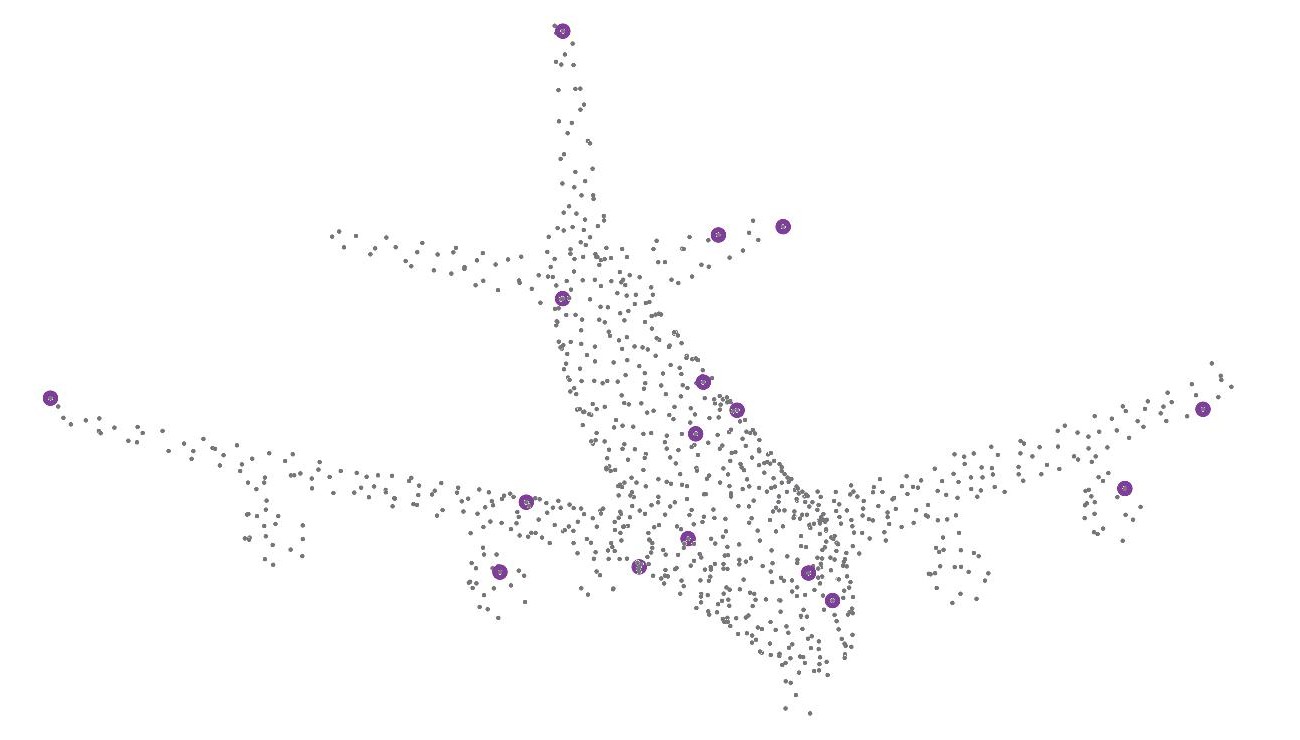}
        \label{fig:16 sampled + surface}
    \end{subfigure}
    \hfill
    \begin{subfigure}[b]{0.245\textwidth}
        \centering
        \includegraphics[width=\linewidth]{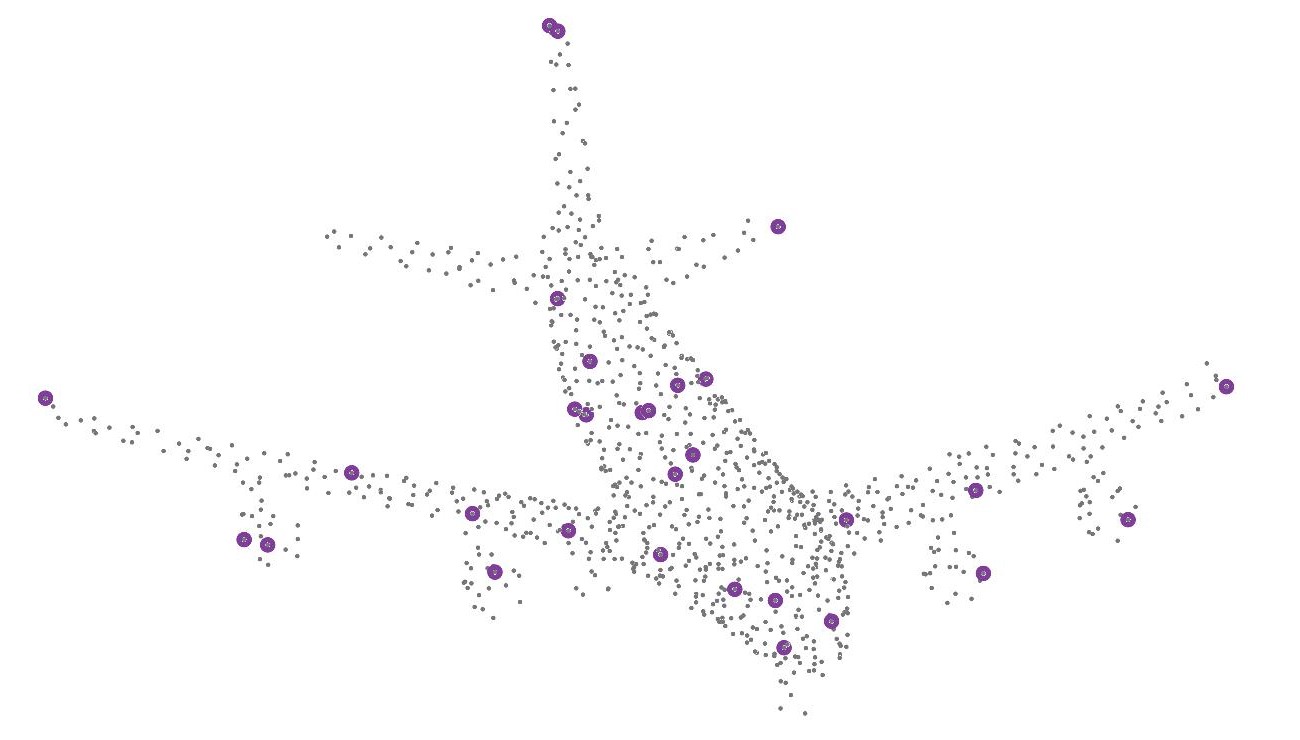}
        \label{fig:32 sampled + surface}
    \end{subfigure}
    \hfill
    \begin{subfigure}[b]{0.245\textwidth}
        \centering
        \includegraphics[width=\linewidth]{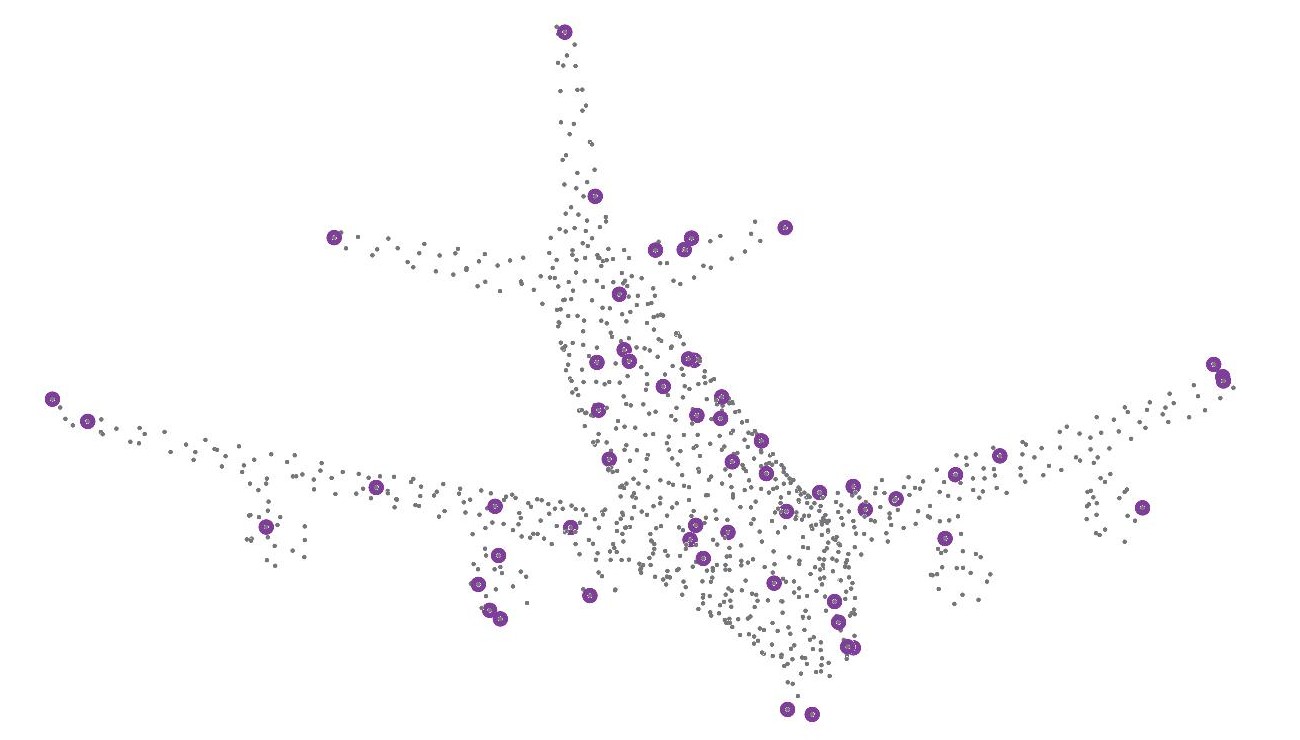}
        \label{fig:64 sampled + surface}
    \end{subfigure}
    \hfill
    \begin{subfigure}[b]{0.245\textwidth}
        \centering
        \includegraphics[width=\linewidth]{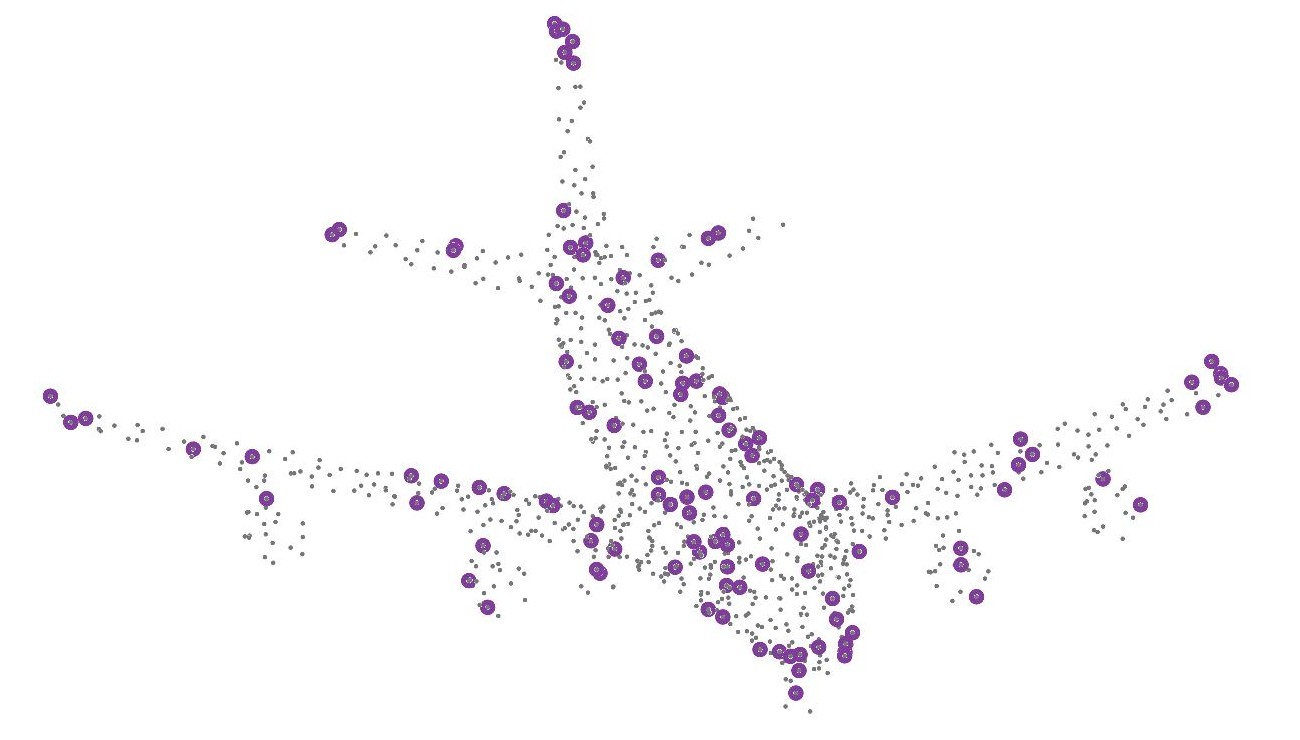}
        \label{fig:128 sampled + surface}
    \end{subfigure}
    \caption{The sampled points through the MorphoSkel3D method, together with the surface points of an airplane example. From left to right the four ratios: 16, 32, 64 and 128 sampled points.}
    \label{fig:sampling example}
\end{figure*}

The pre-trained classification network is optimized for inputs with 1024 surface points. During training, the model underwent data augmentation to improve its generalization and to follow the standard setting for classification. The upper bound classification accuracy is 89.5\% when tested on the same number of 1024 points. To further test its robustness against lower point cloud densities, a comparison is conducted between the traditional FPS and learnable methods. Although our MorphoSkel3D method employs the learnable sampling process from the skeleton-aware (SA) method proposed by Wen et al.~\cite{wen2023learnable}, the objective is to conduct an ablation study that compares the ability of the skeleton as a prior to enhance sampling weights. On the one hand, SA aims to learn the object skeleton in an unsupervised manner similar to P2S~\cite{Lin_2021_CVPR}. On the other hand, MS3D introduces an efficient, object-agnostic approach to find the maximal balls for skeletal extraction. The methods are evaluated across four sampling ratios: 8, 16, 32, and 64 to reduce the original 1024 points to 128, 64, 32, and 16 points respectively. In \cref{tab:objectclassification}, the results confirm that a learnable sampling strategy is significantly more effective than FPS, with the accuracy gap widening as the sampling ratio increases. For a sampling ratio of 8, MS3D maintains the accuracy of the upper bound with a limited decrease at higher ratios. When comparing the influence of the skeleton between SA and MS3D, it can be deducted that the incorporation of a morphological skeleton achieves improvements to better guide the sampling over all four sampling ratios. The advantage is most pronounced at the highest sampling ratio with 85.5\%. The outcome of the learned sampling is demonstrated in \cref{fig:sampling example}, where it becomes apparent that the sampled points are spread across the intricate parts of the object.

\begin{table}[ht!]
\centering
\resizebox{0.47\textwidth}{!}{
\begin{tabular}{c|cccccc}
  \Xhline{1.5pt}
  Ratio & FPS~\cite{qi2017pointnetplusplus} & S-NET~\cite{Dovrat_2019_CVPR} & SN~\cite{lang2020samplenet} & MS~\cite{cheng2022metasampler} & SA~\cite{wen2023learnable} & \cellcolor{gray!20}\MorphoSkel \\
  \Xhline{1.5pt}
  8  & 70.4 & 77.5 & 83.7 & 88.0 & 89.1 & \cellcolor{gray!20}\textbf{89.5} \\
  16 & 46.3 & 70.4 & 82.2 & 85.5 & 88.8 & \cellcolor{gray!20}\textbf{88.9} \\
  32 & 26.3 & 60.6 & 80.1 & 81.5 & 87.4 & \cellcolor{gray!20}\textbf{87.8} \\
  64 & 13.5 & 36.1 & 54.1 & 61.6 & 82.9 & \cellcolor{gray!20}\textbf{85.5} \\
  \Xhline{1.5pt}
\end{tabular}}
\caption{Object classification results on ModelNet40, OA (\%).}
\label{tab:objectclassification}
\end{table}

\subsection{Point Cloud Retrieval}

The integration of morphological skeleton demonstrated an effective selection of points in fine-grained features and, therefore, we transfer the learned sampling networks initially trained for classification to perform shape retrieval. We follow the evaluation scheme of SA~\cite{wen2023learnable}, where the resulting global feature vector is used to search for similar point clouds based on Euclidean distance. In \cref{tab:pointcloudretrieval}, the retrieval results are reported to showcase the additional performance boost of MS3D across all four sampling ratios. 

\begin{table}[ht!]
\centering
\resizebox{0.47\textwidth}{!}{
\begin{tabular}{c|ccccc}
  \Xhline{1.5pt}
  Ratio & FPS~\cite{qi2017pointnetplusplus} & S-NET~\cite{Dovrat_2019_CVPR} & SN~\cite{lang2020samplenet} & SA~\cite{wen2023learnable} & \cellcolor{gray!20}\MorphoSkel \\
  \Xhline{1.5pt}
  8  & 58.3 & 60.4 & 68.8 & 72.2 & \cellcolor{gray!20}\textbf{72.9} \\
  16 & 49.4 & 59.0 & 65.2 & 70.9 & \cellcolor{gray!20}\textbf{71.4} \\
  32 & 37.7 & 59.0 & 62.5 & 67.1 & \cellcolor{gray!20}\textbf{68.7} \\
  64 & 27.4 & 54.5 & 59.5 & 62.6 & \cellcolor{gray!20}\textbf{66.7} \\
  \Xhline{1.5pt}
\end{tabular}}
\caption{Point cloud retrieval results on ModelNet40, mAP (\%).}
\label{tab:pointcloudretrieval}
\end{table}

%% file: sec/6_conclusion.tex
\section{Conclusion} \label{sec:conclusion}
We propose a method that studies a distance function of inner points to highlight the maximal balls in the skeleton. In particular, our approach that applies simple morphological operations on a neighborhood graph is shape-agnostic. The defined skeleton is then utilized to formulate a sampling strategy that emphasizes surface points in intricate regions. This technique enhances the sampling network's ability to learn a representative subset, facilitating two downstream tasks in the representation of the original point cloud. Beyond these two applications, we believe our framework can lay the groundwork to scale prior knowledge in geometric processing for deep learning models.


%% file: sec/7_acknowledgement.tex
\section{Acknowledgement} \label{sec:Acknowledgement}
This work was granted access to the HPC resources of IDRIS under the allocation 2023-AD011014750 made by GENCI.

%% file: sec/X_suppl.tex
\clearpage
\setcounter{page}{1}
\maketitlesupplementary

\begin{table*}[b]
\centering
\resizebox{\textwidth}{!}{%
\begin{tabular}{c|cccc|cccc|cccc|cccc}
\Xhline{1.5pt}
& \multicolumn{4}{c|}{CD-Recon} & \multicolumn{4}{c|}{HD-Recon} & \multicolumn{4}{c|}{CD-MAT} & \multicolumn{4}{c}{HD-MAT} \\
\Xhline{1.5pt}
& P2S & \cellcolor{gray!20}BPA & \cellcolor{gray!20}Alpha & \cellcolor{gray!20}Poisson & P2S & \cellcolor{gray!20}BPA & \cellcolor{gray!20}Alpha & \cellcolor{gray!20}Poisson & P2S & \cellcolor{gray!20}BPA & \cellcolor{gray!20}Alpha & \cellcolor{gray!20}Poisson & P2S & \cellcolor{gray!20}BPA & \cellcolor{gray!20}Alpha & \cellcolor{gray!20}Poisson \\
\Xhline{1.5pt}
Airplane & 0.0363 & \cellcolor{gray!20}0.0187 & \cellcolor{gray!20}0.0158 & \cellcolor{gray!20}\textbf{0.0152} & 0.1266 & \cellcolor{gray!20}0.0911 & \cellcolor{gray!20}\textbf{0.0882} & \cellcolor{gray!20}0.0993 & 0.0611 & \cellcolor{gray!20}0.0547 & \cellcolor{gray!20}0.0431 & \cellcolor{gray!20}\textbf{0.0368} & 0.1721 & \cellcolor{gray!20}\textbf{0.1310} & \cellcolor{gray!20}0.1435 & \cellcolor{gray!20}0.1598 \\
Chair & 0.0441 & \cellcolor{gray!20}0.0362 & \cellcolor{gray!20}0.0365 & \cellcolor{gray!20}\textbf{0.0296} & \textbf{0.1618} & \cellcolor{gray!20}0.1700 & \cellcolor{gray!20}0.2415 & \cellcolor{gray!20}0.2347 & 0.0974 & \cellcolor{gray!20}0.0906 & \cellcolor{gray!20}0.0756 & \cellcolor{gray!20}\textbf{0.0659} & \textbf{0.2151} & \cellcolor{gray!20}0.2241 & \cellcolor{gray!20}0.2906 & \cellcolor{gray!20}0.3123 \\
Table & 0.0424 & \cellcolor{gray!20}0.0369 & \cellcolor{gray!20}\textbf{0.0328} & \cellcolor{gray!20}0.0366 & 0.1745 & \cellcolor{gray!20}\textbf{0.1647} & \cellcolor{gray!20}0.2208 & \cellcolor{gray!20}0.2205 & 0.0876 & \cellcolor{gray!20}0.0745 & \cellcolor{gray!20}\textbf{0.0651} & \cellcolor{gray!20}0.0823 & \textbf{0.2085} & \cellcolor{gray!20}0.2192 & \cellcolor{gray!20}0.2813 & \cellcolor{gray!20}0.3107 \\
Lamp & 0.0335 & \cellcolor{gray!20}0.0233 & \cellcolor{gray!20}\textbf{0.0215} & \cellcolor{gray!20}0.0265 & \textbf{0.1382} & \cellcolor{gray!20}\textbf{0.1382} & \cellcolor{gray!20}0.1491 & \cellcolor{gray!20}0.1896 & 0.0884 & \cellcolor{gray!20}0.0639 & \cellcolor{gray!20}\textbf{0.0575} & \cellcolor{gray!20}0.0606 & \textbf{0.2003} & \cellcolor{gray!20}0.2095 & \cellcolor{gray!20}0.2230 & \cellcolor{gray!20}0.2595 \\
Guitar & 0.0179 & \cellcolor{gray!20}0.0140 & \cellcolor{gray!20}0.0089 & \cellcolor{gray!20}\textbf{0.0085} & 0.0625 & \cellcolor{gray!20}\textbf{0.0486} & \cellcolor{gray!20}0.0490 & \cellcolor{gray!20}0.0553 & 0.0536 & \cellcolor{gray!20}0.0351 & \cellcolor{gray!20}\textbf{0.0238} & \cellcolor{gray!20}0.0259 & 0.1216 & \cellcolor{gray!20}\textbf{0.0864} & \cellcolor{gray!20}0.0992 & \cellcolor{gray!20}0.1046 \\
Earphone & 0.0399 & \cellcolor{gray!20}0.0231 & \cellcolor{gray!20}\textbf{0.0166} & \cellcolor{gray!20}0.0245 & \textbf{0.1125} & \cellcolor{gray!20}0.1502 & \cellcolor{gray!20}0.1859 & \cellcolor{gray!20}0.1686 & 0.1638 & \cellcolor{gray!20}0.1015 & \cellcolor{gray!20}0.1014 & \cellcolor{gray!20}\textbf{0.0934} & \textbf{0.2130} & \cellcolor{gray!20}0.2205 & \cellcolor{gray!20}0.2712 & \cellcolor{gray!20}0.2815 \\
Mug & 0.0417 & \cellcolor{gray!20}\textbf{0.0402} & \cellcolor{gray!20}0.0434 & \cellcolor{gray!20}0.0580 & 0.1419 & \cellcolor{gray!20}0.1439 & \cellcolor{gray!20}\textbf{0.1122} & \cellcolor{gray!20}0.3201 & 0.1179 & \cellcolor{gray!20}0.1171 & \cellcolor{gray!20}\textbf{0.1122} & \cellcolor{gray!20}0.1126 & \textbf{0.2158} & \cellcolor{gray!20}0.2343 & \cellcolor{gray!20}0.3354 & \cellcolor{gray!20}0.4121 \\
Rifle & 0.0213 & \cellcolor{gray!20}0.0133 & \cellcolor{gray!20}0.0107 & \cellcolor{gray!20}\textbf{0.0097} & 0.0767 & \cellcolor{gray!20}\textbf{0.0571} & \cellcolor{gray!20}0.0602 & \cellcolor{gray!20}0.0584 & 0.0356 & \cellcolor{gray!20}0.0353 & \cellcolor{gray!20}0.0263 & \cellcolor{gray!20}\textbf{0.0250} & 0.0957 & \cellcolor{gray!20}\textbf{0.0736} & \cellcolor{gray!20}0.0927 & \cellcolor{gray!20}0.0873 \\
\Xhline{1.5pt}
Average & 0.0372 & \cellcolor{gray!20}0.0294 & \cellcolor{gray!20}\textbf{0.0272} & \cellcolor{gray!20}0.0274 & 0.1424 & \cellcolor{gray!20}\textbf{0.1373} & \cellcolor{gray!20}0.1796 & \cellcolor{gray!20}0.1857 & 0.0828 & \cellcolor{gray!20}0.0714 & \cellcolor{gray!20}\textbf{0.0608} & \cellcolor{gray!20}0.0629 & \textbf{0.1898} & \cellcolor{gray!20}0.1903 & \cellcolor{gray!20}0.2365 & \cellcolor{gray!20}0.2601 \\
\Xhline{1.5pt}
\end{tabular}
}
\caption{Comparison of reconstruction error, for Point2Skeleton as reference and MorphoSkel3D under different surface reconstruction modules, to the surface point cloud (Recon) and the ground truth skeleton (MAT), Chamfer (CD) and Hausdorff (HD) distances.}
\label{tab:reconstructiontraditionalresults}
\end{table*}

\section{Rationale of \MorphoSkel}
\begin{figure}[ht!]
    \centering
    \begin{subfigure}[b]{0.115\textwidth}
        \centering
        \includegraphics[width=\linewidth]{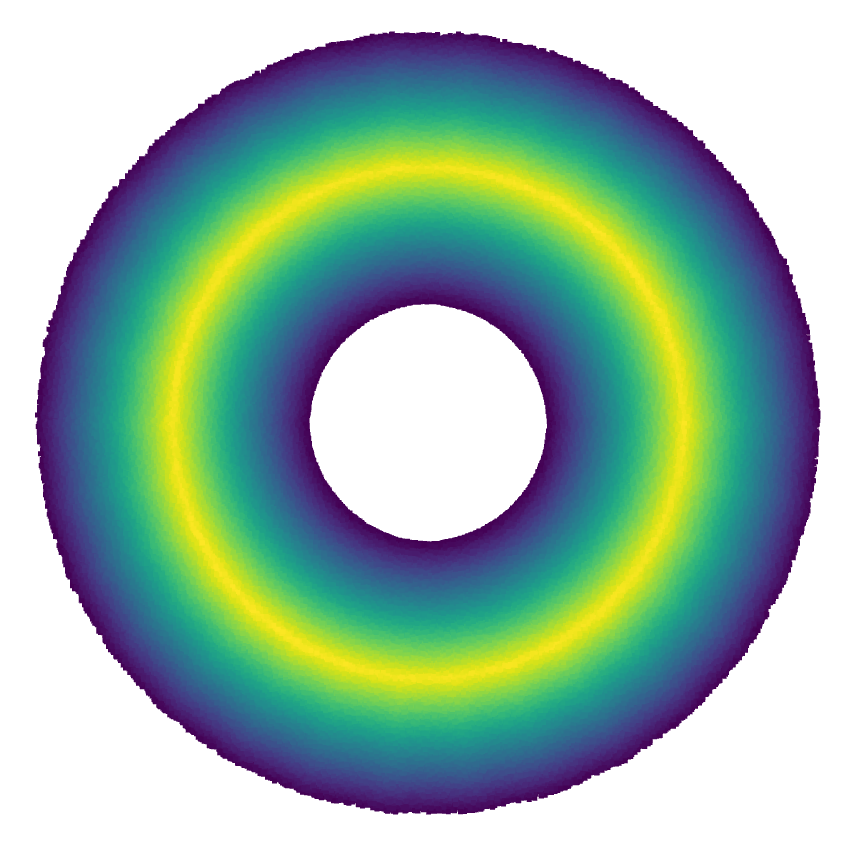}
        \includegraphics[width=\linewidth]{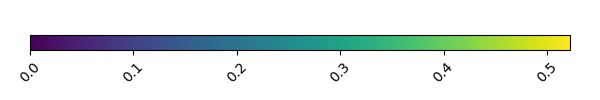}
    \end{subfigure}
    \hfill
    \begin{subfigure}[b]{0.115\textwidth}
        \centering
        \includegraphics[width=\linewidth]{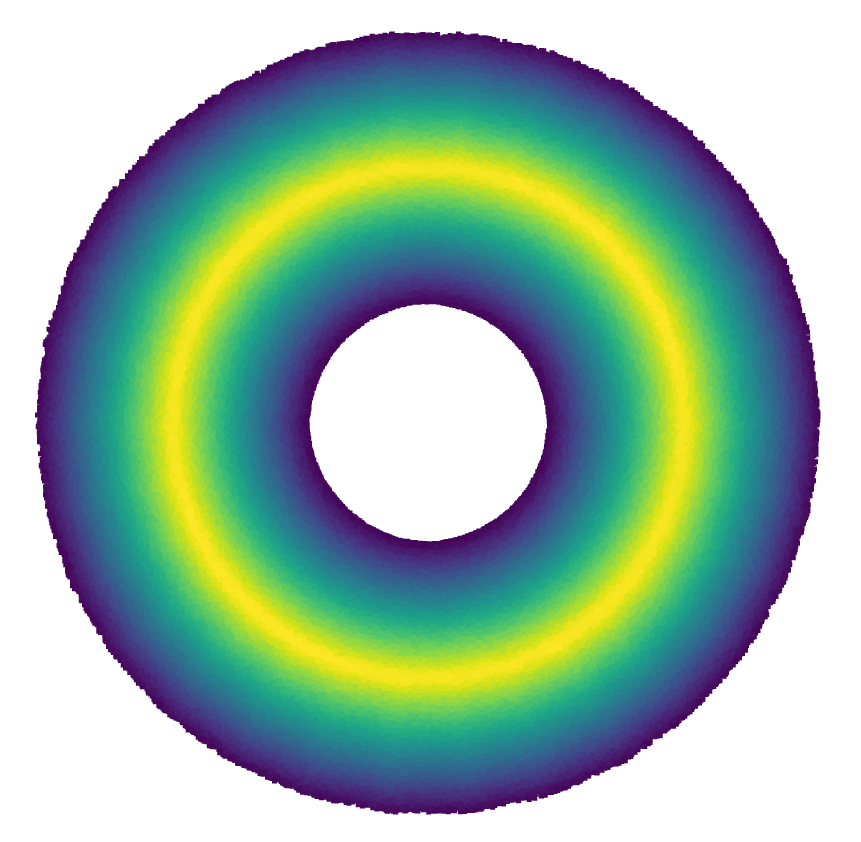}
        \includegraphics[width=\linewidth]{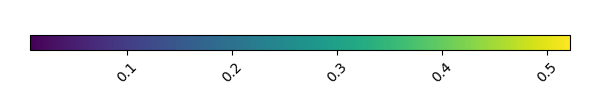}
    \end{subfigure}
    \hfill
    \begin{subfigure}[b]{0.115\textwidth}
        \centering
        \includegraphics[width=\linewidth]{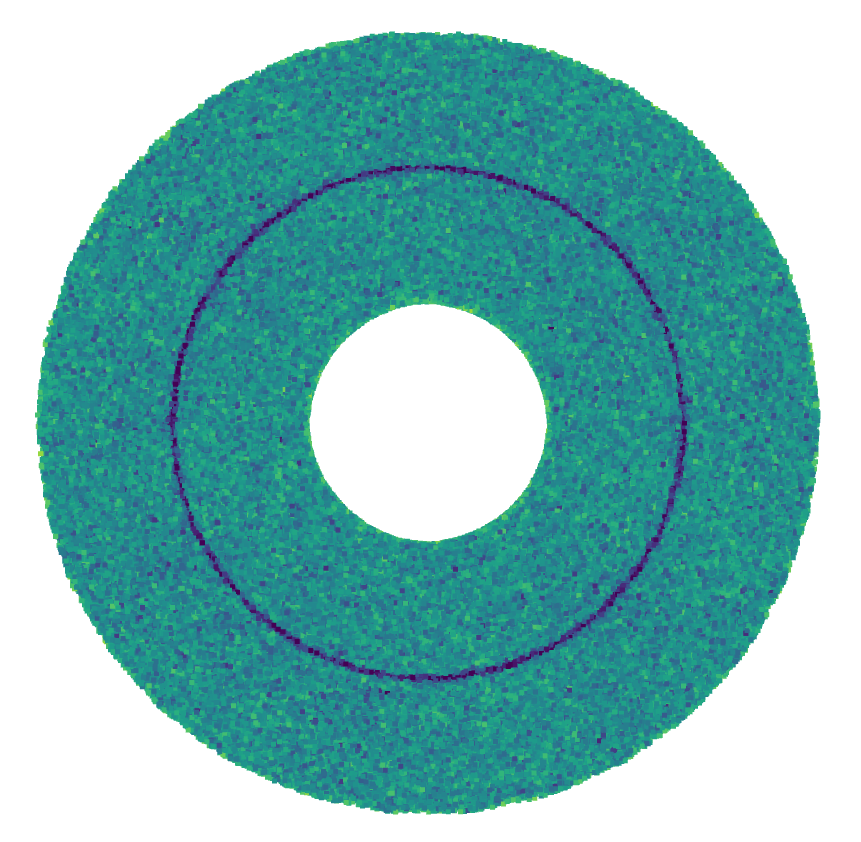}
        \includegraphics[width=\linewidth]{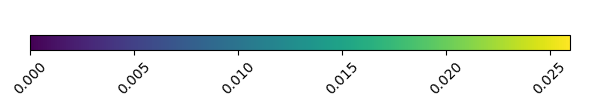}
    \end{subfigure}
    \hfill
    \begin{subfigure}[b]{0.115\textwidth}
        \centering
        \includegraphics[width=\linewidth]{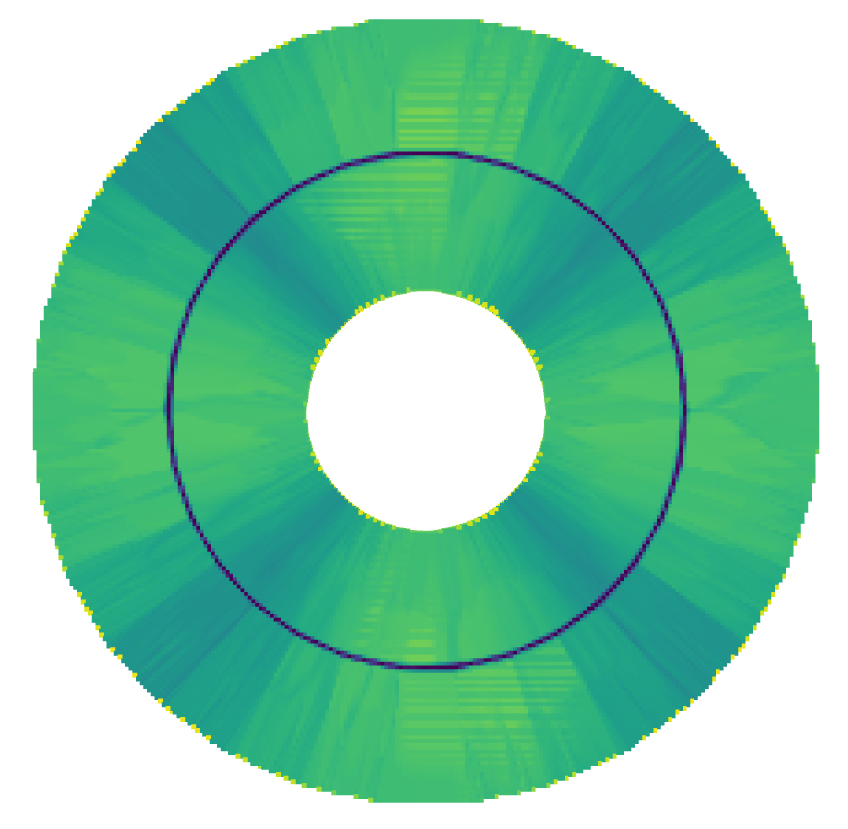}
        \includegraphics[width=\linewidth]{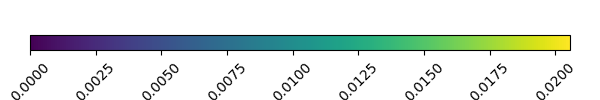}
    \end{subfigure}
    \hfill
    \\[0.25cm]
    \begin{subfigure}[b]{0.115\textwidth}
        \centering
        \includegraphics[width=\linewidth]{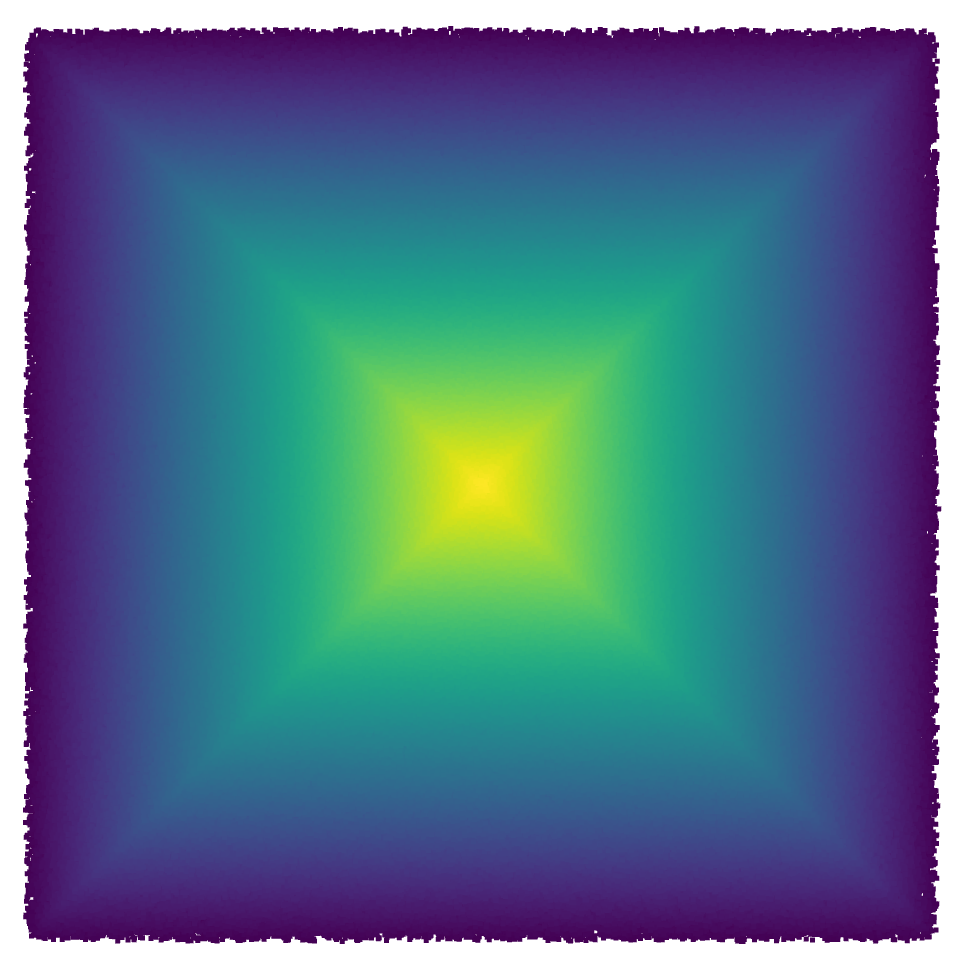}
        \includegraphics[width=\linewidth]{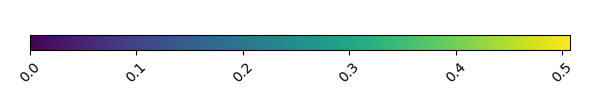}
        \caption{$\text{UDF}(x)$}
        \label{distance function}
    \end{subfigure}
    \hfill
    \begin{subfigure}[b]{0.115\textwidth}
        \centering
        \includegraphics[width=\linewidth]{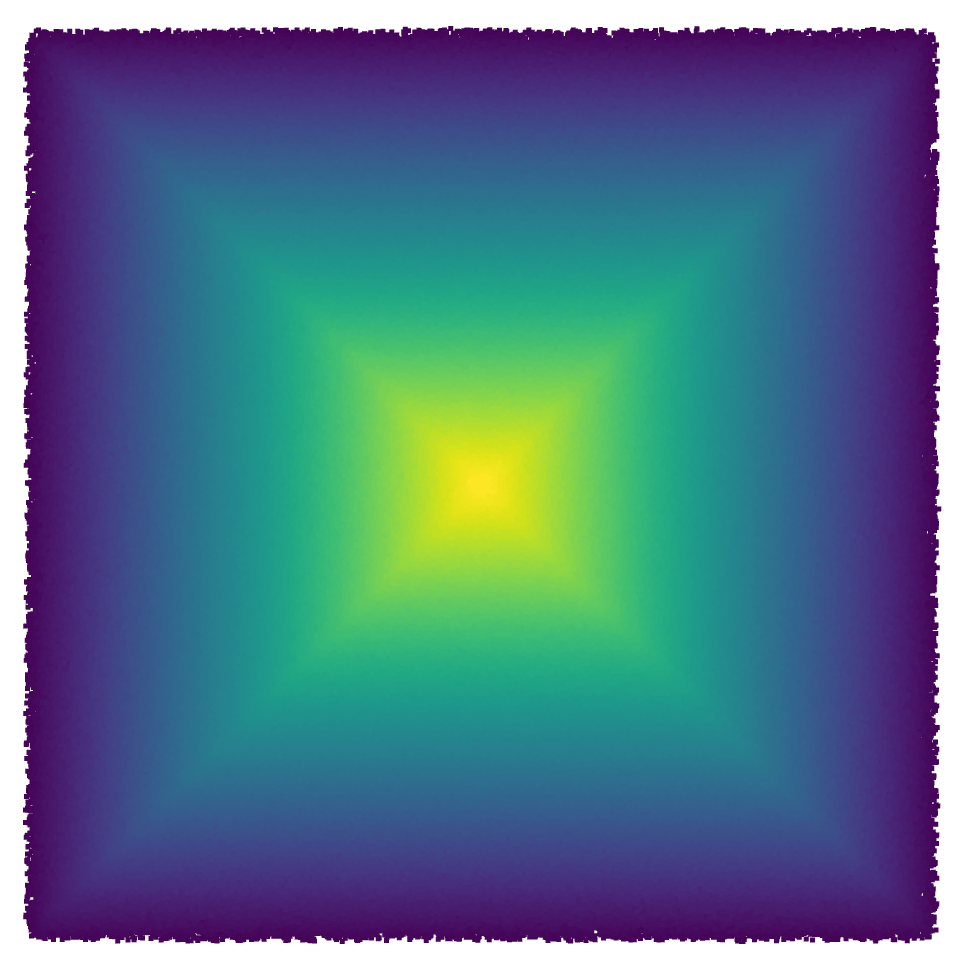}
        \includegraphics[width=\linewidth]{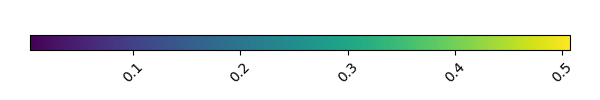}
        \caption{$\delta_{\text{SE}}(\text{UDF}(x))$}
        \label{dilated distance function}
    \end{subfigure}
    \hfill
    \begin{subfigure}[b]{0.115\textwidth}
        \centering
        \includegraphics[width=\linewidth]{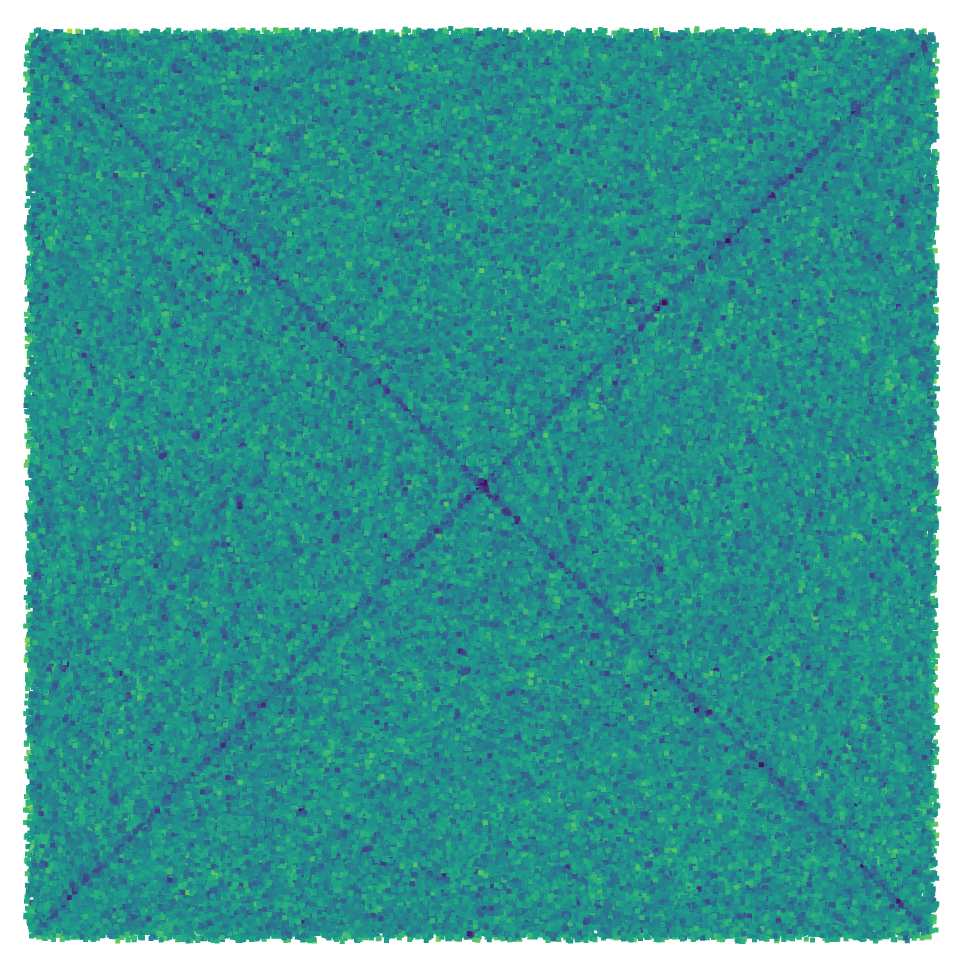}
        \includegraphics[width=\linewidth]{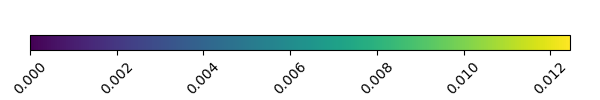}
        \caption{$\MorphoSkel_{\text{rand}}$}
        \label{MS3D random distance function}
    \end{subfigure}
    \hfill
    \begin{subfigure}[b]{0.115\textwidth}
        \centering
        \includegraphics[width=\linewidth]{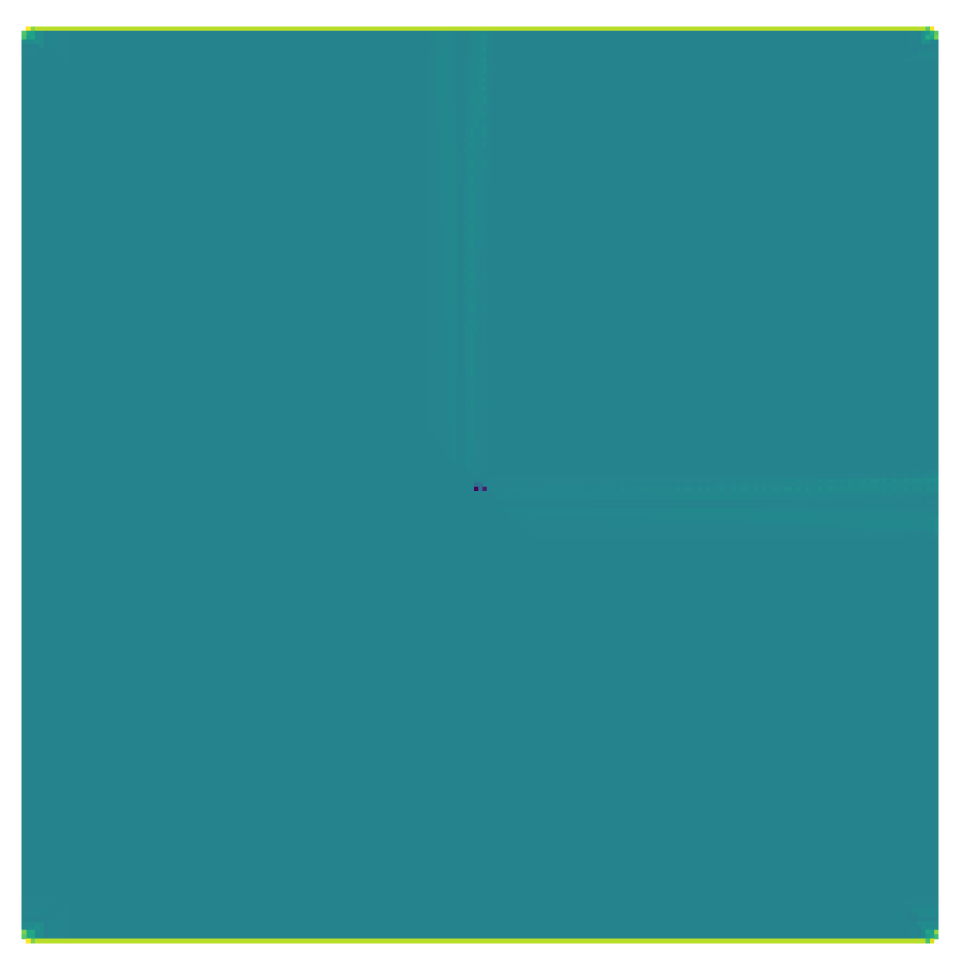}
        \includegraphics[width=\linewidth]{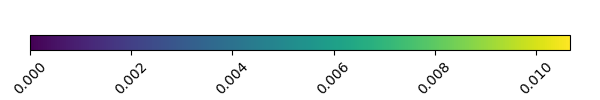}
        \caption{$\MorphoSkel_{\text{grid}}$}
        \label{MS3D meshgrid distance function}
    \end{subfigure}
    \caption{MorphoSkel3D for torus and box with $k$ set to 20. First three columns represent the process for random points, while last column shows the outcome of inner points formed by a meshgrid.}
    \label{fig:rationale}
\end{figure}

The distance function from a set of interior points to the surface is shown in \cref{distance function}, to illustrate how the distance increases as it approaches the geometric center. Using a dilation operator that takes the maximum value of the distance function within the neighborhood defined by a structuring element, we demonstrate in \cref{dilated distance function} that the distance function is expanded to larger values. Except, for the set of maximal balls, the distance remains unchanged from the dilation in \cref{MS3D random distance function} to reveal the skeleton. An alternative to using random inner points is to form a 3D meshgrid that enforces a stricter selection of skeletal points as illustrated in \cref{MS3D meshgrid distance function}.

\vspace{-4pt}
\section{Traditional Surface Reconstruction}
\begin{figure}[ht!]
    \centering
    \begin{subfigure}[b]{0.155\textwidth}
        \centering
        \includegraphics[width=\linewidth]{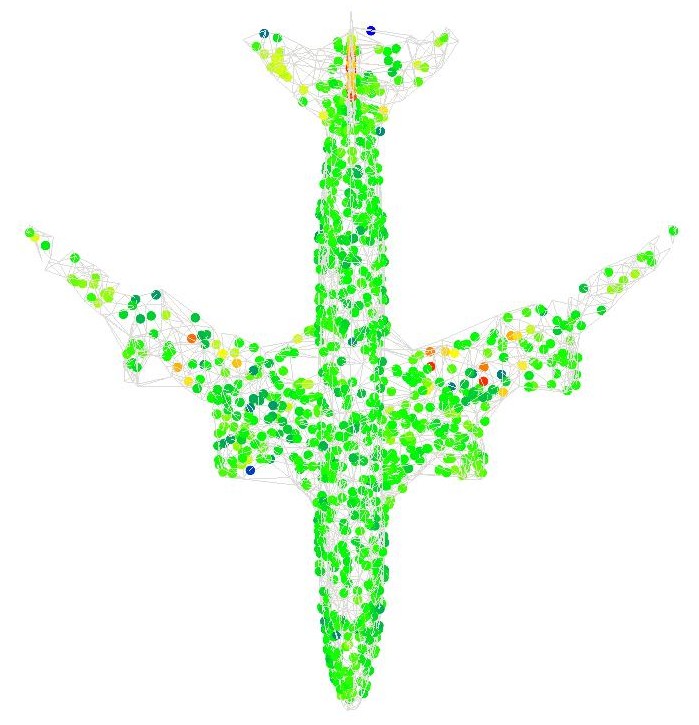}
        \caption{BPA}
        \label{fig:bpa}
    \end{subfigure}
    \hfill
    \begin{subfigure}[b]{0.155\textwidth}
        \centering
        \includegraphics[width=\linewidth]{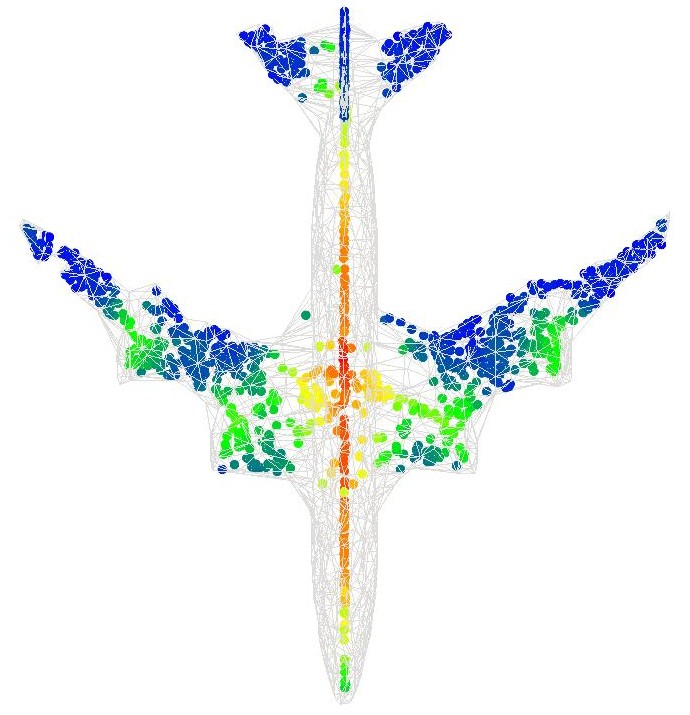}
        \caption{Alpha}
        \label{fig:alpha}
    \end{subfigure}
    \hfill
    \begin{subfigure}[b]{0.155\textwidth}
        \centering
        \includegraphics[width=\linewidth]{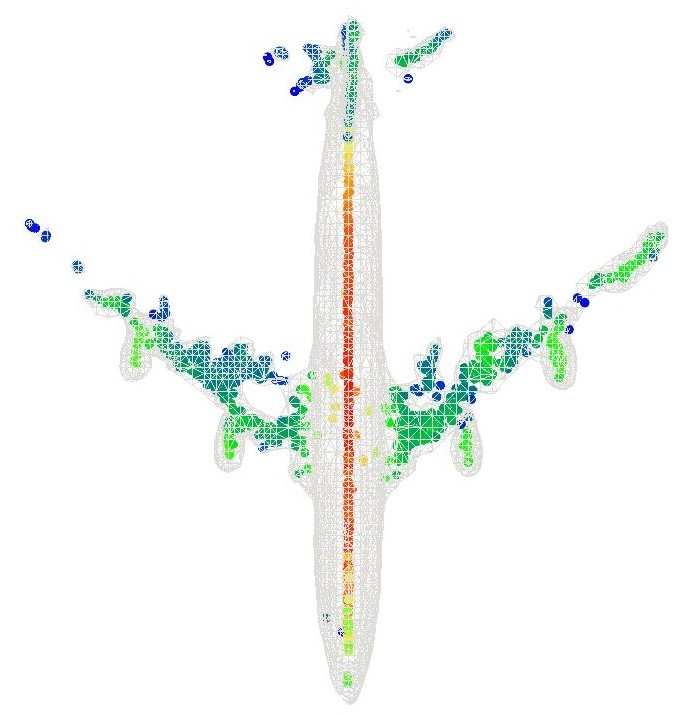}
        \caption{Poisson}
        \label{fig:poisson}
    \end{subfigure}
\caption{The skeletal spheres that are produced from the mesh by three different traditional surface reconstruction algorithms.}
\label{fig:traditionalreconstructionvisualization}
\end{figure}

Given that MorphoSkel3D relies on surface reconstruction within its pipeline, we present a benchmark of different algorithms to evaluate the influence on the skeletal quality. Notably, reconstruction is challenging for the ShapeNet subset with 2000 surface points. In \cref{fig:bpa}, the ball pivoting algorithm (BPA) fails to correctly reconstruct the mesh as the inner points remain on the surface. In contrast, \cref{fig:alpha,fig:poisson} illustrate how alpha shapes and Poisson both create a closed surface to identify inner points for MS3D to skeletonize. In \cref{tab:reconstructiontraditionalresults}, the reconstruction results tend to show lower Chamfer distances for Alpha and Poisson, and lower Hausdorff distances for P2S and BPA.

Therefore, given our uncertainty about whether these metrics fully capture skeletal quality, we also assessed in the experiments how well the skeleton can guide sampling. In \cref{tab:skeletoncomparison}, the classification performance of the learning-to-sample~\cite{wen2023learnable} method is reported for different skeletons.

\begin{table}[ht!]
\centering
\resizebox{0.27\textwidth}{!}{
\begin{tabular}{c|ccc}
  \Xhline{1.5pt}
  Ratio & DPC~\cite{Dpoints15} & SA~\cite{wen2023learnable} & \cellcolor{gray!20}\MorphoSkel \\
  \Xhline{1.5pt}
  8  & 89.1 & 89.1 & \cellcolor{gray!20}\textbf{89.5} \\
  16 & 88.8 & 88.8 & \cellcolor{gray!20}\textbf{88.9} \\
  32 & 87.5 & 87.4 & \cellcolor{gray!20}\textbf{87.8} \\
  64 & 82.8 & 82.9 & \cellcolor{gray!20}\textbf{85.5} \\
  \Xhline{1.5pt}
\end{tabular}}
\caption{Object classification results on ModelNet40, OA (\%).}
\label{tab:skeletoncomparison}
\end{table}

\clearpage
\section{Weakly Supervised Part Segmentation}
\begin{table*}[t]
\centering
\resizebox{\textwidth}{!}{\begin{tabular}{cc|c|cccccccccccccccc}
  \Xhline{1.5pt}
  Method & Ratio & Mean & \underline{Air} & Bag & Cap & \underline{Car} & \underline{Cha} & Ear & \underline{Gui} & Kni & \underline{Lam} & Lap & Mot & Mug & Pis & Roc & Ska & \underline{Tab} \\
  \Xhline{1.5pt}
  \multirow{1}{*}{} 
  & 1 & 83.7 & 83.4 & 78.7 & 82.5 & 74.9 & 89.6 & 73.0 & 91.5 & 85.9 & 80.8 & 95.3 & 65.2 & 93.0 & 81.2 & 57.9 & 72.8 & 80.6 \\
  \Xhline{1.5pt}
  \multirow{4}{*}{FPS \cite{qi2017pointnetplusplus}} 
  & 16 & \textbf{80.1} & \textbf{80.1} & 67.8 & 66.0 & \textbf{65.7} & \textbf{88.0} & 64.8 & \textbf{88.3} & 79.9 & 76.4 & \textbf{95.4} & 51.1 & \textbf{89.6} & \textbf{78.2} & 45.6 & \textbf{74.4} & 77.3 \\
  & 32 & \textbf{79.6} & \textbf{78.9} & \textbf{77.5} & \textbf{74.3} & 63.1 & \textbf{87.8} & 58.5 & 88.5 & \textbf{82.7} & 76.3 & \textbf{95.3} & \textbf{60.6} & \textbf{85.4} & \textbf{80.6} & \textbf{48.3} & \textbf{70.6} & 75.8 \\
  & 64 & \textbf{77.3} & \textbf{75.3} & \textbf{69.6} & \textbf{57.6} & \textbf{56.2} & \textbf{85.8} & 60.2 & 86.8 & \textbf{80.8} & 72.5 & \textbf{94.6} & \textbf{60.3} & 77.1 & \textbf{81.1} & \textbf{50.9} & \textbf{67.6} & 75.1 \\
  & 128 & 70.1 & \textbf{73.3} & \textbf{74.8} & \textbf{50.8} & 46.4 & \textbf{77.3} & 46.5 & \textbf{89.1} & \textbf{82.9} & 72.9 & \textbf{94.4} & \textbf{47.3} & \textbf{86.5} & 71.4 & \textbf{48.0} & 57.8 & 61.2 \\
  \hline
  \cellcolor{gray!20} & \cellcolor{gray!20}16 & \cellcolor{gray!20}79.4 & \cellcolor{gray!20}76.8 & \cellcolor{gray!20}\textbf{74.2} & \cellcolor{gray!20}\textbf{70.2} & \cellcolor{gray!20}63.9 & \cellcolor{gray!20}85.3 & \cellcolor{gray!20}\textbf{66.4} & \cellcolor{gray!20}87.9 & \cellcolor{gray!20}\textbf{82.4} & \cellcolor{gray!20}\textbf{77.2} & \cellcolor{gray!20}94.2 & \cellcolor{gray!20}\textbf{57.4} & \cellcolor{gray!20}82.2 & \cellcolor{gray!20}77.3 & \cellcolor{gray!20}\textbf{52.6} & \cellcolor{gray!20}65.8 & \cellcolor{gray!20}\textbf{78.7} \\
  \cellcolor{gray!20} & \cellcolor{gray!20}32 & \cellcolor{gray!20}77.5 & \cellcolor{gray!20}73.0 & \cellcolor{gray!20}74.5 & \cellcolor{gray!20}47.5 & \cellcolor{gray!20}\textbf{64.8} & \cellcolor{gray!20}82.5 & \cellcolor{gray!20}\textbf{59.7} & \cellcolor{gray!20}\textbf{88.6} & \cellcolor{gray!20}80.9 & \cellcolor{gray!20}\textbf{77.9} & \cellcolor{gray!20}95.1 & \cellcolor{gray!20}52.7 & \cellcolor{gray!20}74.8 & \cellcolor{gray!20}73.4 & \cellcolor{gray!20}45.0 & \cellcolor{gray!20}63.8 & \cellcolor{gray!20}\textbf{76.6} \\
  \cellcolor{gray!20} & \cellcolor{gray!20}64 & \cellcolor{gray!20}76.9 & \cellcolor{gray!20}72.3 & \cellcolor{gray!20}69.5 & \cellcolor{gray!20}48.8 & \cellcolor{gray!20}55.8 & \cellcolor{gray!20}84.7 & \cellcolor{gray!20}\textbf{66.2} & \cellcolor{gray!20}\textbf{87.8} & \cellcolor{gray!20}78.2 & \cellcolor{gray!20}\textbf{76.3} & \cellcolor{gray!20}92.4 & \cellcolor{gray!20}49.0 & \cellcolor{gray!20}\textbf{81.3} & \cellcolor{gray!20}80.4 & \cellcolor{gray!20}48.1 & \cellcolor{gray!20}65.5 & \cellcolor{gray!20}\textbf{75.5} \\
  \cellcolor{gray!20}\multirow{-4}{*}{\MorphoSkel} & \cellcolor{gray!20}128 & \cellcolor{gray!20}\textbf{73.3} & \cellcolor{gray!20}68.0 & \cellcolor{gray!20}70.8 & \cellcolor{gray!20}40.7 & \cellcolor{gray!20}\textbf{53.1} & \cellcolor{gray!20}76.9 & \cellcolor{gray!20}\textbf{65.4} & \cellcolor{gray!20}87.8 & \cellcolor{gray!20}80.9 & \cellcolor{gray!20}\textbf{75.5} & \cellcolor{gray!20}93.4 & \cellcolor{gray!20}40.9 & \cellcolor{gray!20}83.3 & \cellcolor{gray!20}\textbf{77.5} & \cellcolor{gray!20}47.4 & \cellcolor{gray!20}\textbf{65.6} & \cellcolor{gray!20}\textbf{72.6} \\
  \Xhline{1.5pt}
\end{tabular}}
\caption{Weakly supervised part segmentation results on different sampling ratios, mIoU (\%).}
\label{tab:shapenet}
\end{table*}

\paragraph{Dataset}
MorphoSkel3D demonstrated effective point sampling across intricate regions and, therefore, we transfer the learned sampling networks initially trained for classification to perform weakly supervised segmentation on ShapeNet in this section. The ShapeNet dataset includes 16 object categories with a total of 50 classes to train a single model for coarse-level segmentation, six of these categories also represented in ModelNet. In specific, the ShapeNet part segmentation dataset contains 2048 points per object where each is labeled with an annotated ground truth. To enhance segmentation performance with prior knowledge, we leverage the pre-trained sampling network from ModelNet to the downstream task of ShapeNet. The sampling network of ModelNet serves as a backbone to train a segmentation model on ShapeNet. It's important to note that a sampled subset is thus provided to learn segmentation, making the segmentation task weakly supervised since not all points and labels are used. The idea is that a pre-trained sampling network of ModelNet selects a representative subset of points and could be transferred to ShapeNet with the same goal. 

\paragraph{Metric}
The evaluation scheme for part segmentation aligns with state-of-the-art, where the intersection over union (IoU) for a shape is derived from the average IoUs of its parts. For each category, the IoU is in turn calculated as the average of all shape IoUs within the category. Finally, the overall instance average mIoU is determined by averaging the IoUs of all instances in the test set. For a fair comparison, we employ no data augmentation techniques to follow the standard setting for part segmentation. Therefore, the fully supervised segmentation results in the upper part of Tab. \ref{tab:shapenet} are identical to those reported in PointNet \cite{qi2016pointnet}. With an instance-averaged mIoU of 83.7\%, an upper bound is established to benchmark weakly supervised methods. In our goal to reduce the annotation effort and effectively learn from a limited set of partially labeled points, we compare the sampling of the classic FPS with MS3D. Both methods are evaluated across the four sampling ratios to reduce the original set into subsets of 128, 64, 32, and 16 points.

\begin{figure}[ht!]
    \centering
    \begin{subfigure}[b]{0.23\textwidth}
        \centering
        \includegraphics[width=\linewidth]{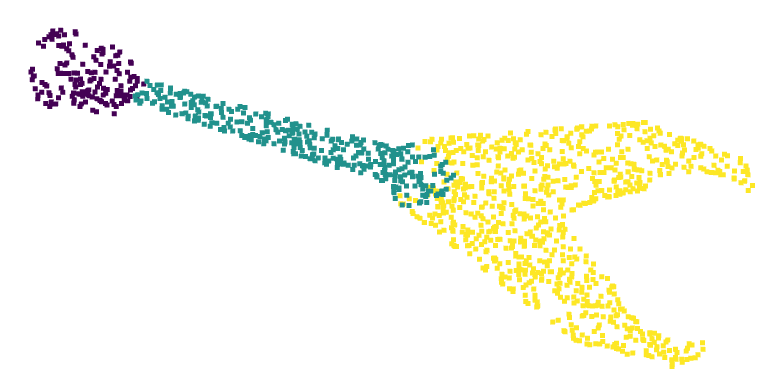}
    \end{subfigure}
    \hfill
    \begin{subfigure}[b]{0.23\textwidth}
        \centering
        \includegraphics[width=\linewidth]{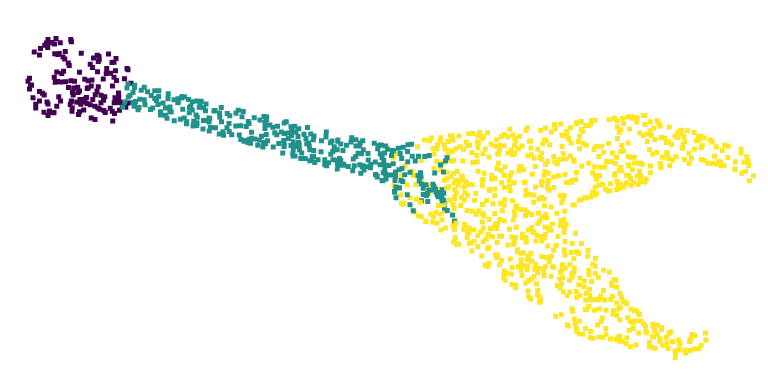}
    \end{subfigure}
    \begin{subfigure}[b]{0.23\textwidth}
        \centering
        \includegraphics[width=\linewidth]{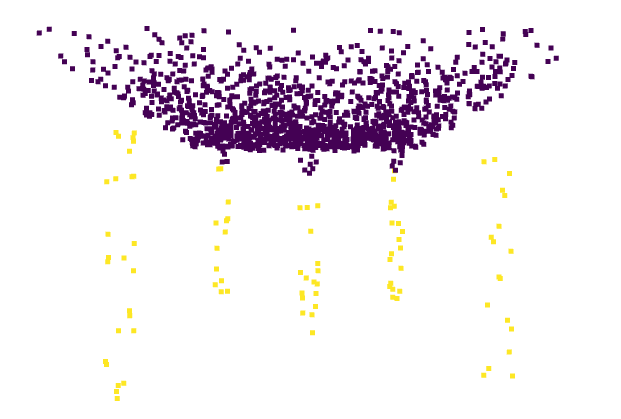}
        \caption{Prediction}
    \end{subfigure}
    \hfill
    \begin{subfigure}[b]{0.23\textwidth}
        \centering
        \includegraphics[width=\linewidth]{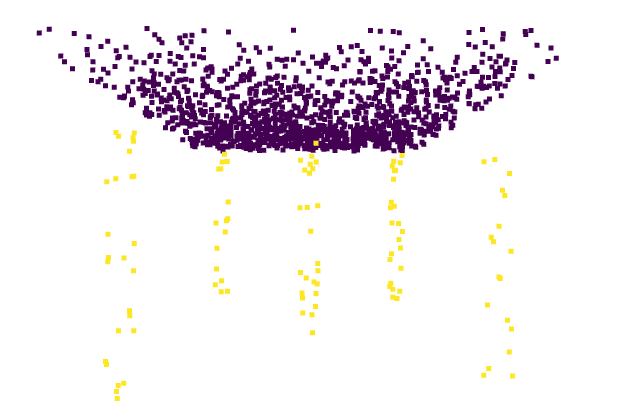}
        \caption{Ground truth}
    \end{subfigure}
    \caption{Segmentation model trained with 16 labeled points, sampled through MorphoSkel3D. The prediction for a guitar and table is displayed on the left, while the ground truth is on the right.}
    \label{fig:segmentation visualization}
\end{figure}

\paragraph{Results}
In Tab. \ref{tab:shapenet}, the performance of FPS is compared against MS3D across 16 categories, with six of these categories underlined as they are also found in ModelNet. To provide more insight in the effectiveness of MS3D, we should focus on, but not limit the analysis to the six categories that occur in ModelNet. At the highest sampling ratio of 128, MS3D indicates an improvement in overall mIoU that surpasses FPS by over 3\% with 73.3\% compared to 70.1\%. For instance, the gap in the table category is evident based on the sampling strategy used by MS3D. It namely targets the corner areas where there's a transition in label between the surface and legs of a table. A segmented example of the table and guitar category is shown in Fig. \ref{fig:segmentation visualization} to compare prediction and ground truth. For the three other sampling ratios, the overall mIoU slightly differs in the favor of FPS. We illustrated that our method focused its sampling in learned regions for classification and, therefore, assume that it fails to cover the entire object as effectively as FPS to learn segmentation. On the other hand, MS3D distributes its few sampled points in different parts. This observation suggests that our proposed method efficiently identifies points in each part to annotate and learn a segmentation. Consequently, the study of objects in a fine-grained segmentation setting arises as an interesting task.

